\documentclass[default,pdflatex,sn-mathphys,iicol]{sn-jnl}% Default with double column layout

\jyear{2022}%

\theoremstyle{thmstyleone}%
\theoremstyle{thmstyletwo}%

\theoremstyle{thmstylethree}%

\usepackage{setspace}
\usepackage{amsmath,amssymb,amsfonts}

\usepackage{bbding}

\usepackage{booktabs}
\usepackage{colortbl}

\usepackage{graphicx}
\usepackage{multirow}
\usepackage{booktabs}
\usepackage{subfigure}
\usepackage{hyperref}
\hypersetup{
hidelinks
}

\begin{document}

\title[Article Title]{A Personalized Utterance Style (PUS) based Dialogue Strategy for Efficient Service Requirement Elicitation}


\author[1]{\fnm{Demin} \sur{Yu}}\email{deminyu98@gmail.com}

\author[1]{\fnm{Min} \sur{Liu}}\email{minmiu77@qq.com}

\author[1]{\fnm{Zhongjie} \sur{Wang* }}\email{rainy@hit.edu.cn*}

\affil[1]{Faculty of Computing, Harbin Institute of Technology, China}

% \affil[2]{\orgdiv{Department}, \orgname{Organization}, \orgaddress{\street{Street}, \city{City}, \postcode{10587}, \state{State}, \country{Country}}}


\abstract{% 随着互联网上众多服务的出现，对于服务提供者来说准确交付服务的前提是如何有效地获取完整准确的用户需求——服务需求获取。通常，用户更喜欢用自然语言表达需求，这种交互方式对于服务提供者来说是低效率的。因此，借助对话系统利用智能交互的方式获取用户需求是重要的。由于用户需求通常包含不同层次的子意图，并且需要被来自不同领域的服务所满足，对于SRE任务存在一个巨大的潜在意图空间来探索完整的需求。考虑到使用静态槽位的传统对话系统无法直接被应用到SRE任务中，使用有效的对话策略引导用户在巨大潜在意图空间中，表达他们的完整准确的意图是一项挑战。用户倾向于按照顺序主观表达需求的倾向的现象，我们提出PUS模型来识别个性化的需求表达习惯，并且应用PUs到一个有效的对话策略来完成SRE任务。具体的，对话测试选择合适的响应动作来动态更新对话状态。借助从对话历史中抽取的PUS，系统将收缩在潜在意图空间中的搜索范围。实验结果表明，具备pus的对话策略能够用更少的对话轮次获得更加准确的用户需求

With the flourish of services on the Internet, a prerequisite for service providers to precisely deliver services to their customers is to capture user requirements comprehensively, accurately, and efficiently. This is called the ``Service Requirement Elicitation (SRE)'' task. Considering the amount of customers is huge, it is an inefficient way for service providers to interact with each user by face-to-face dialog. Therefore, to elicit user requirements with the assistance of virtual intelligent assistants has become a mainstream way. Since user requirements generally consist of different levels of details and need to be satisfied by services from multiple domains, there is a huge potential requirement space for SRE to explore to elicit complete requirements. Considering that traditional dialogue system with static slots cannot be directly applied to the SRE task, it is a challenge to design an efficient dialogue strategy to guide users to express their complete and accurate requirements in such a huge potential requirement space. Based on the phenomenon that users tend to express requirements subjectively in a sequential manner, we propose a Personalized Utterance Style (PUS) module to perceive the personalized requirement expression habits, and then apply PUS to an dialogue strategy to efficiently complete the SRE task. Specifically, the dialogue strategy chooses suitable response actions for dynamically updating the dialogue state. With the assistance of PUS extracted from dialogue history, the system can shrink the search scope of potential requirement space. Experiment results show that the dialogue strategy with PUS can elicit more accurate user requirements with fewer dialogue rounds.}




\maketitle

\section{Introduction}

% 随着计算服务化的发展，越来越多的虚拟服务被部署到互联网中。虽然丰富类别的服务丰富了人们的生活，他也为用户选择随时服务以及服务提供者满足用户的多种需求带来了困难，导致服务供需双方不匹配。准确匹配需求的前提是获得用户的完整需求一遍与服务提供者能够提供定制化的服务组合，并且用户也可以不必过多关注不同类型服务的选择。在这种情况下，对话系统的便携性对未来服务需求获取的发展的需求导向的系统带来了遍历。对话系统能够通过和用户交互的方式辅助服务提供者获得用户需求。在大部分场景下，任务型对话系统能够根据预设的模板或者规则并且主要应用填槽技术，展开多轮对话来完成特殊的任务。然而作为服务计算的重要一部分，如何获得完整需求对于对话兄台你个来说是一项必要但是经常被忽略的任务。
% 从任务型对话系统的角度，我们将SRE作为一项特殊的任务。对于SRE，用户提出的需求经常是有层次的并且跨领域的。一些需求可以被分解为多个细粒度的子需求并且被分布到不同的领域，这对于对话系统就形成了一个硕大潜在意图空间去搜索。如图1所示的例子，不同层次的意图间存在依赖关系。一个完成的意图，如去北京旅行，经常会包含着多个领域，如没事、景点、住宿等。由于用户会动态的暴露他们的意图，SRE可以被视为潜在意图在广阔空间中不断被探索的过程。旅客在想旅行时，经常想订机票并且寻找美食餐厅，然后对餐厅，他们会想去了解当地的特色菜。对话系统的责任就是去有效的引导用户表达他们的完整意图，利用多轮对话动作构成的对话策略。然而，对于传统的包含静态且局限的槽位构成的对话系统，在广大的潜在意图空间中发现目标意图不能被直接解决。
% 本文中，我们介绍了意图树来建模复杂需求结构。我们注意到用户不表达需求的过程并不是随机的，而是倾向于从一个中心向周围发散，并且表达的孙旭有一种确定的主观倾向。提高对话效率的关键就在于利用引导用户可能意图，以及基于当前上下文和用户状态引导意图表述，来缩小搜索空间。考虑到用户再表达意图时存在个性化的倾向，我们提出PUS模型来从对话历史中学习用户意图表达的倾向。当前有很多工作关注用户的个性化特征，如文本风格或者用户画像偏好，然而，这些工作只是聚焦于在个人属性偏好的变化，没有考虑表达意图层面的偏好， 值得注意的是用户在计划需求表达时有一个主观的倾向，可能会采取个性化的对话行动策略 以确认不同的意图约束。如图所示，当确认旅馆意图时，对话策略需要找到下一个可能的意图并且采取措施引导用户根据他们的个性化偏好。因此，PUS从三个层面描述用户：意图规划的偏好，动作状态转移的皮那好以及需求属性的偏好。IP关注于用户following intention的可能趋势；AST关注用户在多轮对话中采取的不同对话动作的个性化特点；RA类似与用户画像，记录用户对意图约束的偏好。
% 对于意图树模型，我们提出特殊的对话策略实现SRE任务。具体的，我们介绍了一种需求模式，一种频繁模式的需求组合表示方式，来引导对话方向。对话策略利用需求模式和PUS选择合适的需求来动态拓展对话状态中的意图树，来以引导用户表达模糊的意图。我们在定制化的CrossWOZ数据集上应用我们的对话来测试对话效果。结果显示融合pus的对话策略能够获得更好的结果

The Service Requirement Elicitation (SRE) has been becoming an important task in service oriented computing. With the development of computing servitization, more and more virtual services are deployed on the web.
The variety of services that enriches life, also makes it difficult to choose appropriate services that meet user's personalized requirements.
% While the variety of services enriches people's life, it also brings difficulties for users to choose the most appropriate services according to their requirements. 
Without complete and accurate requirements, service providers cannot support requirements with customized services, thus leading to a mismatch between service supply and demand. The prerequisite for service providers is to capture diverse user requirements with efficient method, which is called the ``Service Requirement Elicitation (SRE)'' task. 
% The complete and accurate requirements is the key to decoupling users and service providers. 
% Users can pay less attention to the choice of various services, and service provider can precisely deliver customized services based on requirements. 
However it is an inefficient way for service providers to communicate with customers by face-to-face dialog considering the huge amount of customers. In this case, it is becoming a trend to elicit user requirements with the assistance  of intelligent dialogue systems~\cite{yu2021incorporating,tian2022intention}.
\begin{figure}[ht]
\centering
\includegraphics[width=1\columnwidth]{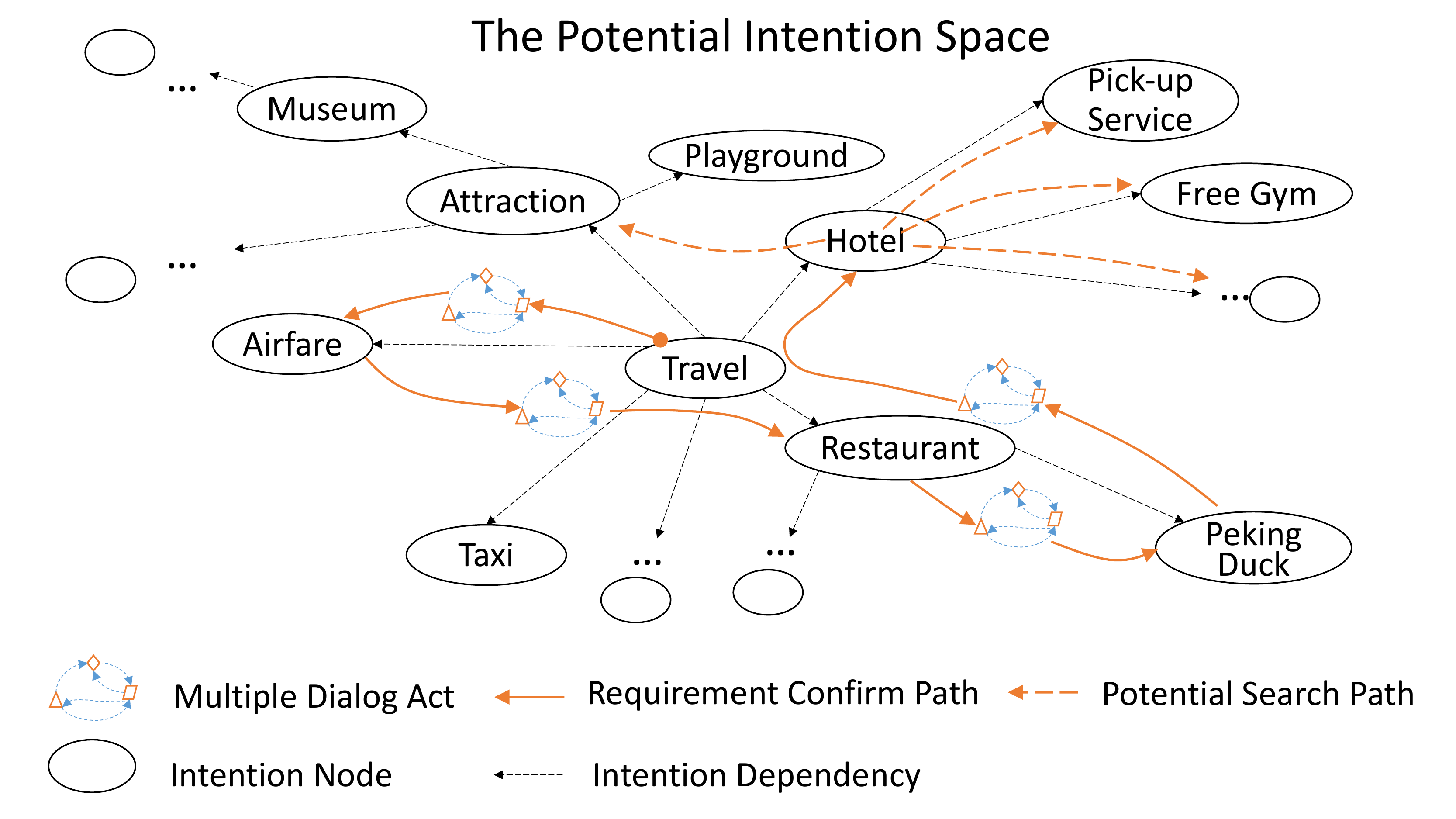}
\caption{The Process of Exploring Requirements in Potential Requirement Space.}
\label{fig:intent_space}
\end{figure}

We adopt a special tree-like structure to model the complex requirement in dialogue. Unlike traditional dialogue tasks like flight checking or accommodation reservation, the dialog system for SRE task pays less attention to the entity recommendation but to capturing accurate needs of various domains ~\cite{xu2020domain}. In this way, the requirements in SRE task enable the decoupling of users and service providers. However, the requirements proposed by users in SRE are generally complex, hierarchical and transboundary. Most complex requirements can be divided into more finer-grained sub-requirements, and generally need to be satisfied by services from different domains. We introduce the Requirement Tree ($R\_Tree$) \cite{tian2022intention} to model complex requirements. From the perspective of dialog system, the huge amount of candidate requirements supported by multi-domain services and the dependencies among different levels of sub-requirements constitute a potential requirement space ~\cite{xu2020domain}.

We propose \textbf{Personalized Utterance Style} (PUS)to describe requirement expression preference to assist requirement elicitation.  Considering that users have personalized habits when expressing their requirements~\cite{mazare2018training}, the PUS characterizes users from three dimensions: Requirement Planning (RP) preference, Act State Transfer (AST) preference, and Requirement Attribute (RA) preference. The RP focuses on the tendency of following potential requirements. The AST aims to figure out features of dialogue actions in multiple rounds. The RA, like user profile, is used to record preference of constraints.

% It is worth noting that the intention and the requirement express a similar concept in SRE task. 
We regard our dialogue strategy as a requirement node searching and optimizing process in a huge potential requirement space. 
% Since users expose their requirements dynamically, the SRE can be regarded as a process of exploring potential requirements in the huge requirement space. 
The goal of dialog system for SRE is to efficiently guide users to express requirements and generate the target $R\_Tree$. As an example of requirement space shown in Fig. \ref{fig:intent_space}, there are dependencies between different levels of requirements distributed in multiple domains. Part of a traveler's requirements is to find a restaurant, and then for the restaurant, they want to learn about the local specialties. Once confirming the requirement about hotel, dialog system needs to guide user to express the following potential requirement, such as attraction or sub-requirements about hotel. 
To elicit complete and accurate service requirements in huge requirement space, we propose a PUS-based dialogue strategy, which uses $R\_Tree$ to model dialog state and target dialog goal. Specifically, our strategy chooses suitable requirements for dynamically expanding the requirement tree of dialogue state with the assistance of PUS. The PUS can assist strategy generates suitable response dialog action with guiding requirement expression by narrowing the scope of potential requirement space. We apply our strategy to customized CrossWOZ dataset \cite{zhu2020crosswoz}, and results show that the strategy which integrates utterance style can elicit more accurate and complete user requirements with fewer dialogue rounds and a higher success rate.

The main contributions of this work are as follows: 
\begin{itemize}
\item We propose  \textbf{Personalized Utterance Style} (PUS) to portray the requirements expression preference from the perspective of the whole personalized dialog context. 
\item We design a PUS-based dialog strategy to elicit complex requirements to generate dialog target $R\_Tree$ by shrinking the search scope in the potential requirement space. 
\item We prove that the PUS-based dialog strategy can effectively elicit more accurate and complete user requirements and complete conversation in limited rounds. Furthermore, the introduced PUS feature can improve the performance of deep neutral network for dialogue.
\end{itemize}

% The remainder of this paper is organized as follows.
% Section \ref{sec:related_work} describes related studies in task-oriented dialog systems and personalized features in dialogue.
% Section \ref{sec:problem} introduces problem formulation of SRE task.
% Section \ref{sec:method} proposes our methodology of PUS modeling and dialog strategy for the SRE task.
% Section \ref{sec:experiment} introduces experiments and results.
% Section \ref{sec:conclusion} presents our conclusion and future work.

\section{Related Work}
% 添加 Service requirement 相关研究

\label{sec:related_work}
\subsection{Task-oriented Dialogue System for Service Requirement} 
Service requirements elicitation plays a decisive role in the service oriented computing ~\cite{lee2019structural}. 
Xiang et al.~\cite{xiang2007srem} propose to use ontology list to to narrow generic service requirements and capture services requirements.
Hwang et al. ~\cite{hwang2014service} regard QoS as discrete random variables with probability mass functions to address the service selection problem.
The goal of requirement elicitation task is to find out complete requirements as much detail as possible so that it could become easier to support downstream task~\cite{wang2021graph}.  
%如何准确有效地感知用户需求在服务领域是一项重要的任务。
% How to accurately and effectively perceive user requirements is an important task in the service sector. Traditional methods use user records to identify intentions. Kaiya\cite{Kaiya2006Using} used ontology to build a domain knowledge base for requirement analysis.
However, these approaches strongly rely on domain data and manual rules, leading to bad interaction of users. In this paper, we consider the SRE task as a special dialogue task.

Conversation is a user-friendly method to elicit user requirements.
% 然而这些方法依赖于领域数据和人工规则，并且存在糟糕的用户交互。
With the development of dialogue technology in recent years, more and more task-oriented dialogue systems are applied to interact with users to provide specific services. 
There are many challenges for dialogue system to achieve specific tasks.
% 近些年随着对话技术的发展，越来越多的任务型对话系统被用来与用户交互来向用户提供具体的服务。
First of all, the dialogue study has been blocked by the scale of data available. Zhu et al. \cite{zhu2020crosswoz} propose a multi-domain dialogue dataset with rich annotation within travel scenarios to advance dialogue system modeling.
% 首先，对话研究被缺少有效数据所阻碍。zhu等人提出了在旅行场景下具有充分标注的多领域对话数据集来促进对话系统的建模。
Secondly, depending on the different functions of dialogue systems, research usually focuses on dialogue state tracking (DST) and dialogue policy (DP) to improve dialogue effectiveness.
% 其次，根据对话系统的不同功能，研究通常聚焦于对话状态追踪和对话策略两个层面提升对话效果。
Lee et al. \cite{lee2019sumbt} propose a universal and scalable model called SUMBT to track slots state by learning the relations between domain-slot-types and slot-values appearing in utterances.
Takanobu et al. \cite{takanobu2019guided} propose a join policy optimization and reward estimation method to estimate the reward signal and infers the user goal in the dialog sessions.
Zhang et al. \cite{zhang2020task} propose an three stage end-to-end dialogue model to handle the problem that multiple responses can be appropriate for the same dialog context.
However, these dialogue systems with static slots of specific tasks and are limited to cover complete requirements and dialogue state, resulting in an inability to handle the user's complex requirements and continue the dialog.
% 然而，这些对话系统的对话内容都依赖于特定任务的对话数据，使用固定的槽位描述用户的意图以及系统对话状态，导致无法有效处理用户的复杂需求
% 然而这些方法可解释性差，对数据有强依赖性，我们尝试通过对话的方式引导用户表达其潜在需求。

\subsection{Personalized Features in Dialogue}
\begin{table}[htbp]
\small
\centering
\caption{Different task of personalized dialogue}
\label{table:persona}
\begin{tabular}{ccc} 
\toprule
\textbf{Task}                                                                  & \begin{tabular}[c]{@{}c@{}}\textbf{Related }\\\textbf{Work}\end{tabular} & \begin{tabular}[c]{@{}c@{}}\textbf{Personalized }\\\textbf{Information}\end{tabular}            \\ 
\hline
\begin{tabular}[c]{@{}c@{}}Personalized\\~Dialogue \\Consistency~\end{tabular} & \cite{qian2021learning}  \cite{madotto-etal-2019-personalizing}~                                                               & \begin{tabular}[c]{@{}c@{}}Language Style\\~\& User Profiles\end{tabular}                       \\ 
\hline
\begin{tabular}[c]{@{}c@{}}Personalized\\Response~\\Generation\end{tabular}    & \cite{zhang2020memory} \cite{luo2019learning}~~                                                              & \begin{tabular}[c]{@{}c@{}}User Profiles \&\\~Knowledge\\~Template\end{tabular}                 \\ 
\hline
\begin{tabular}[c]{@{}c@{}}Personalized \\Recommendation\end{tabular}          & \cite{he2021conversation}~                                                               & \begin{tabular}[c]{@{}c@{}}Dialogue History, \\User Profiles \&\\~Knowledge~ ~ ~~\end{tabular}  \\
\bottomrule
\end{tabular}
\end{table}

Creating a virtual personal assistants (VPA) that can have realistic conversations with humans is one of the ultimate goals of Artificial Intelligence (AI). 
One of the challenges of VPA is to take the personalized information into account in order to achieve a customized conversation service for the user.
% VPA的挑战之一是考虑用户个性化信息，以实现为用户提供定制化的对话服务。
% Hence, it is very important to take the personalized information into account
Current research about user personalization are distributed in different stage of dialogue system.
% 当前针对用户个性化的研究点分布在对话过程的不同部分。
% 从对用户建模的不用角度分析：利用用户信息作推荐；个性化对话语言生成；用户性格的针对性对话策略。
Qian et al. \cite{qian2021learning} propose the IMPChat that owns the consistent personality with the corresponding user by modeling implicit user profile. %\cite{madotto-etal-2019-personalizing}
% modeling user’s personalized language style and personalized preferences separately.
Zhang et al. \cite{zhang2020memory} achieve the personalized response retrieval and generation based on similar user profile attributes. %\cite{luo2019learning}
He et al. \cite{he2021conversation} propose to harness dialog historical information to adapt different scenarios and leverage the knowledge base and user profiles to achieve high quality conversational recommendation.
We summarize the relevant works on personalized information for different dialogue tasks, as shown in Table \ref{table:persona}.
% 我们总结了不同对话任务对个性化信息的相关研究，如表所示
However, these studies have focused more on users' profile attributes for external entities, ignoring the users' preference of dialogue utterance from the whole context.
% 然而这些研究都更加关注在用户对外部实体的偏好，忽略了用户自身在对话表达过程中的习惯。
%The main goal of dialogue system research is to achieve fully natural and human-like human-computer interaction with different users. Based on this, many works focus on the study of user cognition. First, people try to consider user information in dialogue recommendation. Second, many works focus on the correspondence of dialogue strategies to individual dialogue user information. In addition, some works try to generate linguistic responses that conform to their characteristic style according to different users.
% However, these modeling of user characteristics ignore the influence of the user's personalization tendency in dialogue habits on dialogue strategies.
% 对话系统研究的主要目标是与不同用户都能够实现充分自然的、类人的人机交互，这要求对话系统除了需要具备有效的知识推理、对话策略之外，还应该能够对用户又足够且充分的认知。基于此，许多工作聚焦于对用户认知的研究。首先，人们试图在对话推荐中考虑用户信息。其次，很多工作关注对话策略对个性话用户信息的相应。最后，一些工作试图根据不同的用户生成符合其特征风格的语言回复。
% 然而，这些对用户特征的建模都忽视了用户对话习惯上的个性化倾向对对话策略的影响。
Unlike the traditional user profile, the personalized utterance style (PUS) in this work aims to characterize the user’s requirement expression preference during the conversation. Considering that language style or user profile can be distinguished or extracted from discrete utterances, the PUS requires the whole dialogue context to be analyzed.
% 概括地讲，文本风格或者用户profile能够从离散的对话语句中分辨或者被抽取，而对话风格需要从完整的对话上下文中分析。

% Intention recognition  系统理解用户意图
% Requirement elicitation 系统获取用户需求

\section{Problem Formulation}
\label{sec:problem}
% \my{a seperate section problem formulation... as in this paper, you introduce too much new terms. After read Section IV, the framework section is the problem formulation section?}
\subsection{Requirement Model for SRE}
\label{sec:task}
% \my{It seems the focuss of this subsection is talking about I-tree, rather than SRE...}
% 
% ==========
%% Demin : Mainly introduce the $R\_Tree$ model, concept of ``intention'' and ``requirement''; is suitable to include dialog system and concept of dialog for SRE in there

\begin{figure}[ht]
\centering
\includegraphics[width=\linewidth]{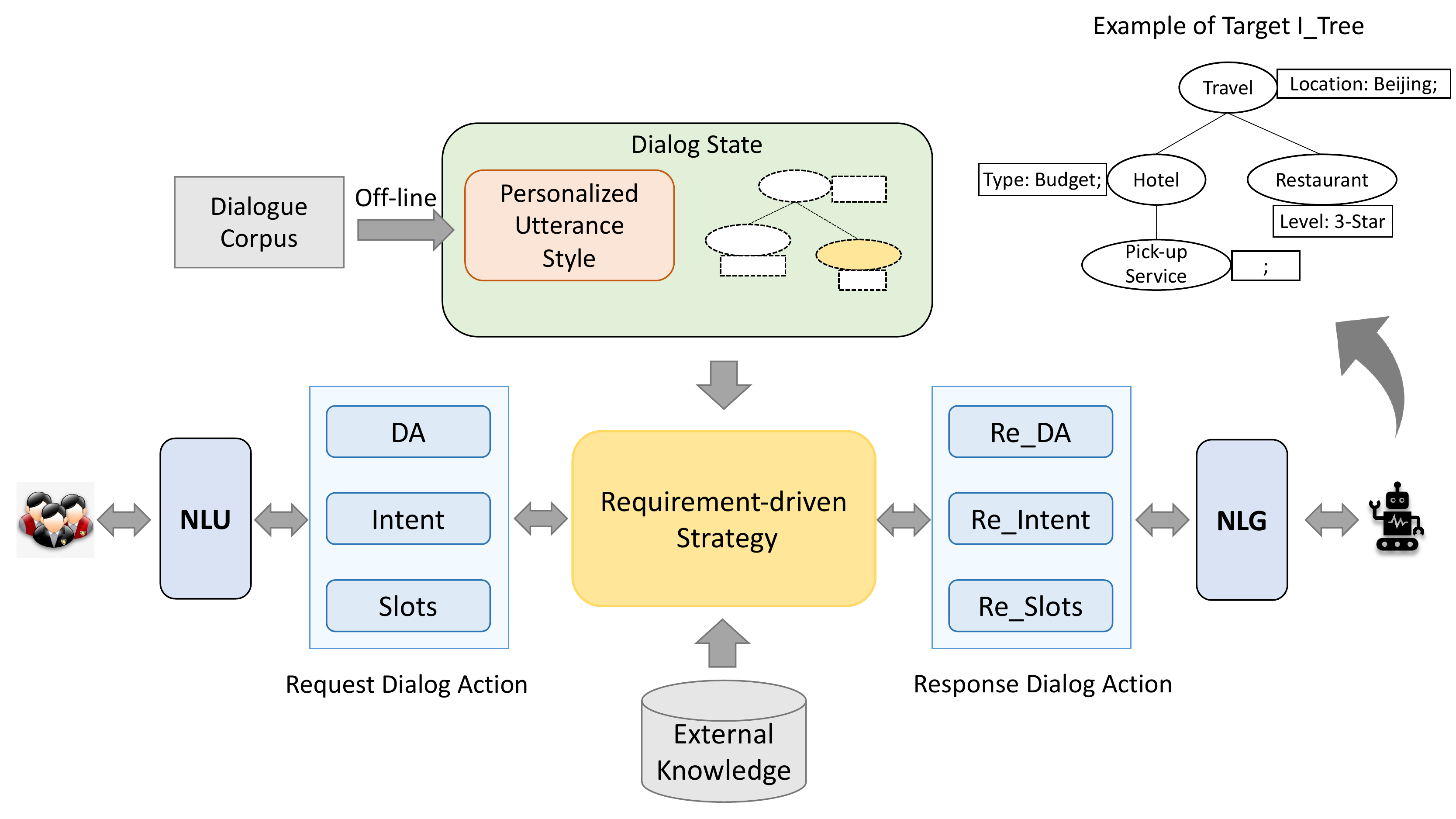}
\caption{Overview of our Dialogue Framework }
% \my{highlight your work.}}
\label{fig:framework}
\end{figure}

We introduce the requirement tree ($R\_Tree$)~\cite{tian2022intention} to model the complex structure of user requirements in SRE task. 
% 不同于常规任务型对话，% 本文研究的对话系统将引导用户表达完整需求作为对话目标。与常规对话推荐相比，需求获取任务主要关注对话用户主观需求。由于现实中实体状态是不断变化的，对话机器人不关注需求的作为下游任务的实体推荐。我们使用前文提出的需求树模型来描述用户需求。
The $R\_Tree$ can be considered as a tree structure consisting of Requirement Nodes. The definition of $R\_Tree$ is shown in the following equation:
\begin{equation}
    R\_Tree = (V,E)
\end{equation}
Where $V$ is the set of requirement nodes and $E$ is the edges between requirements, which describes the dependency between two nodes. 
The edge $e_{ij} \in E$ stands for the directed edge from $v_i$ to $v_j$, which means that $v_i$ is the sub-requirement of $v_j$.
Each requirement node $v_i \in V$ can be described as following equation:
\begin{equation}
    v_i = (Req, Slots, Sub\_Reqs)
\end{equation}
Where $Req$ describes user's functional requirements, which is textual expression. $Sub\_Reqs$ donates the dependencies of child nodes. For requirement node $v_i$, the child nodes can be defined as $Sub\_Reqs_i = \{v_j \vert \exists e_{ij} \in E\}$. 
$Slots$ represents the non-functional constraints of user requirements. 
% \my{the distance of fig and context is too large..}. 
For each $Slot \in Slots$, it is defined as equation~\ref{eq:slots}.
\begin{equation}
    \label{eq:slots}
    Slot = (Slot\_Name, Slot\_Value)
\end{equation}
Where $Slot\_Name$ and $Slot\_Value$ denote the specific attribute of constraint.
The target $R\_Tree$ in Fig. \ref{fig:framework} shows an example of decomposing user requirements into an requirement tree.

As we introduce dialog system to elicit service requirements, the goal of system is to generate the target $R\_Tree$ by multiple rounds of dialog.
For every requirement expressed by user, dialog system construct details of requirement as an requirement node and update the $R\_Tree$ based on dependencies among requirements.
% Typically, the ``requirement'' and ``intention'' express the same concept from the different perspectives of user and system.

% \my{for problem definition, what you supposed to do is formally describe this problem (such as math.,), not just use some toy examples.. toy examples are used in introduction. So, actually, you still dont tell me what SRE is.}
% As for the SRE task, the dialogue system needs to interact with user with multiple rounds and elicit whole user requirements, which are set in advance of conversation.
% The user will express their pre-set requirements, such as ``I want to travel to Beijing''. Then the dialogue system should take actions to guide the user to express left requirements in as few turns as possible. Finally, the dialogue system generates the target $R\_Tree$ of user requirement by the information in dialogue context.
% The goal of SRE task is to match the target $R\_Tree$ elicited by system with the initial user $R\_Tree$ as closely as possible.
% 给定用户初始意图树，用户向对话系统表达顶层需求以开启对话，系统通过与用户的多轮对话交互，借助对用户个性化特征的理解，引导用户不断补充当前意图。最终，系统获得对用户完整意图的理解，并以意图树的形式给出。对话目标是系统理解的对话意图树与给定用户的初始意图树尽可能匹配。

\subsection{Framework of Dialogue System}

Considering that completing specific tasks with the assistance of intelligent dialog system has become a mainstream way, we design a dialog system for SRE task based on generic structure of task-oriented dialog system.
The overview of our dialogue system framework is shown in Fig. \ref{fig:framework}, which consists of four major modules: Natural Language Understanding (NLU), Natural Language Generation (NLG), Dialogue State Tracking (DST) and Dialogue Policy (DP)~\cite{chen2017survey}. These modules form a message pipeline to handle user request and generate response. Different with general task-oriented dialogue system, we use requirement tree ($R\_Tree$) rather than static slots to model dialogue state.
We abstract the semantic dialog action into the following triples:
% 我们将用语义理解结果抽象为如下三元组
\begin{equation}
    \label{eq:understand}
    <DA, Req, Slots>
\end{equation}
where $DA$ donates the Dialog Act, which can be interpreted as the atomic units of a conversation \cite{10.1162/tacl_a_00420}. 
Similar to the concept in equation \ref{eq:slots}, the $Slots$ in dialog action donates the different constraints of $Req$.

During the conversation, the dialog policy uses the requirement discussed in the current context as a node pointer to locate the corresponding requirement node in the current dialog state. Based on the request dialogue action and the semantic slot information, relevant operations are performed on the requirement nodes. When the current requirement has been discussed completely, the dialog policy searches the potential requirement space and selects possible other requirements to continue the dialogue to achieve effective guidance. At this time, the requirement tree in the dialog state is also continuously expanded to maintain the current dialog context information. In order to confirm the user's current requirement and guide the expansion of the requirement more efficiently, the dialog policy will combine the user's requirement, the dialog state and the PUS to confirm the appropriate response action.

Note that how to tag user actions from user speech (NLU) or generate neutral language according to dialogue action (NLG) is not our main focus. We try to apply the existing accurate methods, like BERT\footnote{https://github.com/thu-coai/ConvLab-2/tree/master/convlab2/nlu/jointBERT}, to achieve these tasks and do not make enough innovative contributions.

\begin{table*}
\centering
\caption{The Dialogue Acts for Requirement Elicitation Scenarios}
\label{table:act}
\begin{tabular}{llll} 
\toprule
\multicolumn{1}{c}{\textbf{Rough Category}} & \multicolumn{1}{c}{\textbf{Meaning}}                                                            & \multicolumn{1}{c}{\textbf{Detailed Act}} & \multicolumn{1}{c}{\textbf{Example}}  \\ 
\hline
\multirow{2}{*}{Statement}                  & \multirow{2}{*}{\begin{tabular}[c]{@{}l@{}}Offer requirements, \\entity or events\end{tabular}} & State\_in                                 & ~I want to find a five-star hotel.    \\
                                            &                                                                                                 & State\_out                                & The hotel provides pick-up service.   \\ 
\hline
\multirow{3}{*}{Question}                   & \multirow{3}{*}{Offer different questions.}                                                     & Ques\_select                              & Do you prefer A or B?                 \\
                                            &                                                                                                 & Ques\_rec                                 & How about traveling by bus?           \\
                                            &                                                                                                 & Ques\_req                                 & What do you need for the hotel?       \\ 
\hline
\multirow{3}{*}{Response}                   & \multirow{3}{*}{\begin{tabular}[c]{@{}l@{}}Response to \\statements or questions.\end{tabular}} & Resp\_acc                                 & This restaurant is a good choice.     \\
                                            &                                                                                                 & Resp\_deny                                & I don't think so.                     \\
                                            &                                                                                                 & Resp\_vag                                 & I have no opinion.                    \\ 
\hline
General                                     & \begin{tabular}[c]{@{}l@{}}Verbalized acts \\or unknown acts.\end{tabular}                      & General                                   & Thanks; Bye;                          \\
\bottomrule
\end{tabular}
\end{table*}

\subsection{Personalized Utterance Style Models}
\label{sec:pus_model}

\begin{figure}[ht]
\includegraphics[width=\linewidth]{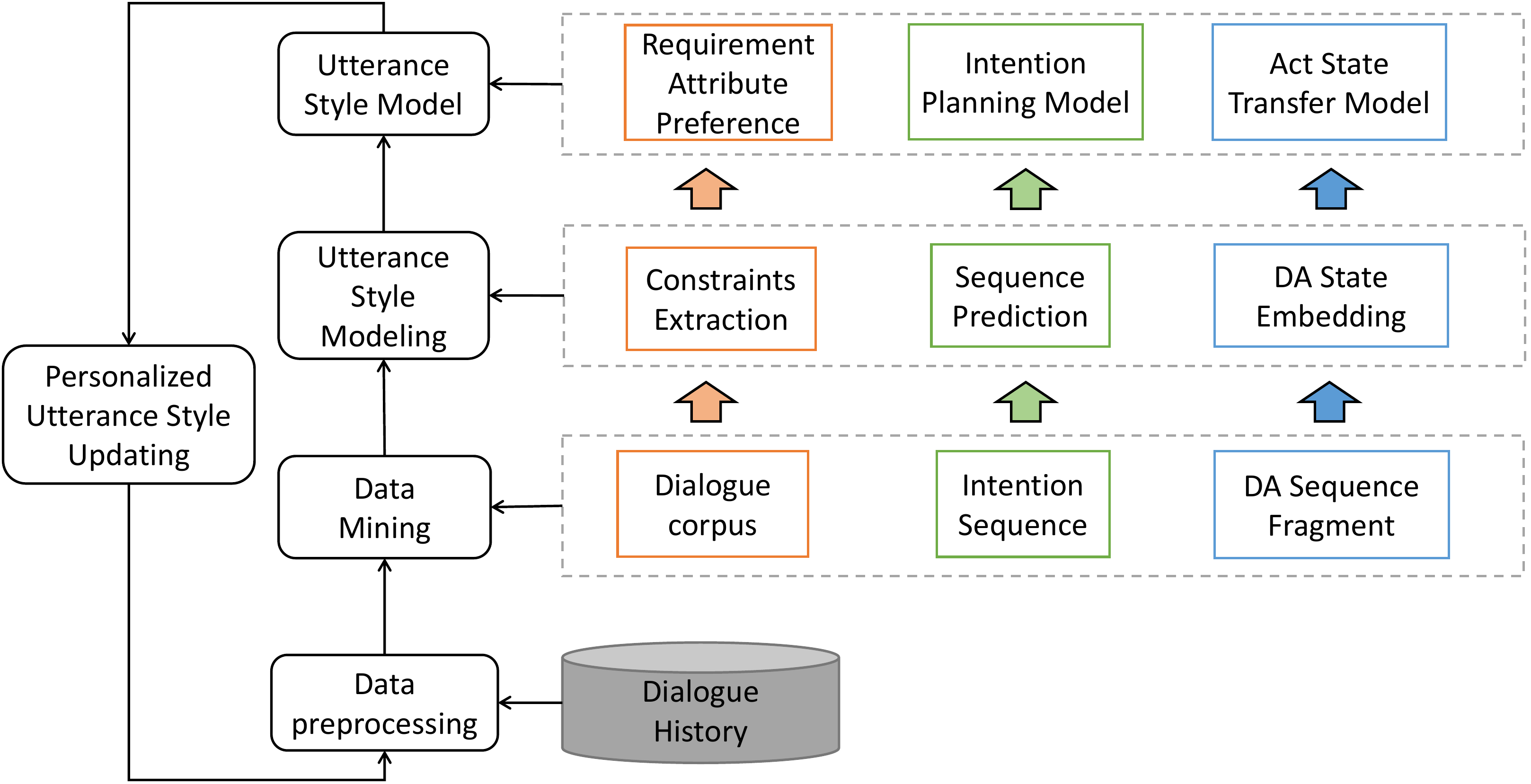}
\caption{The Overview of Personalized Utterance Style Modeling}
\label{fig:style_modeling}
\end{figure}
% 为了让对话系统具备更加准确地感知用户意图的能力，除了对语义信息的理解，对用户对话习惯的把握也是至关重要的。用户个性化对话风格能够帮助对话系统估计用户的需求表达趋势，依据用户对话习惯规划有针对性的对话策略，以更好的引导需求表达

In order for the dialogue system to perceive requirements more accurately, the grasp of the user's dialogue habits is crucial in addition to the understanding of semantic information. Personalized utterance style can help to estimate the user's requirement expression trend and select targeted dialog strategy based on the user's dialog habits, so as to better guide the requirement expression.
In this paper, we propose a \textbf{Personalized Utterance Style} (PUS) module as shown in Fig. \ref{fig:style_modeling}, which mainly consists of three different parts: Act State Transfer (AST) Preference, Requirement Planning (RP) Preference and Requirement Attribute (RA) Preference.

\subsubsection{Act State Transfer Preference} 
Dialog Act ($DA$) represents the semantic motivation of current utterance. 
As an important part of the user requirement expression, learning user dialog state state transfer preferences is an important task.
% 作为用户表达个人意图的重要部分，对用户对话动作选择偏好的认知是一项重要的任务。
We focus on modeling act state transfer habits from dialog acts sequence in dialogue history. %Specifically, we use weighted finite state transducers (FST) 
The goal of this part is to learn $DA$ states to generate system response act according to the user request act. The act state transfer preference, modeled as a state transfer graph, can be described as $A=(Q,\Sigma,\delta)$, where $Q$ donates the all act state trained from of DA sequence. The $\Sigma$ donates the set of all dialog acts. The $\delta$ donates the state transfer functions defined as follows:
\begin{equation}
    \label{eq:act_transfer}
    DA_{out} = \delta(s,DA_{in})
\end{equation}
where $DA \in \Sigma$ donates the input and output dialog acts, and the $s \in Q$ donates the current state. 
% 值得注意的是，不同对话场景包含了不同的对话动作体系。针对SRE任务，我们结合现有的，设计了覆盖对话意图的动作行为。
According to related linguistic theory \cite{zhou2010casia} \cite{bunt2010towards}, we design a specific dialogue act category for the SRE task shown on Table. \ref{table:act}.
% 根据语言学理论，我们为需求获取任务设计了有针对性的对话行为体系
Some cases in scenarios of SRE show that the DA system in Table \ref{table:act} can basically cover the different utterances of dialogue.

\subsubsection{Requirement Planning Preference} The requirement planing model represents the user preference of requirement expression during the dialogue, which is able to guide the order of dialogue topic.
The requirement planning preference can be modeled as Equation \ref{eq:intent_plan}
% \my{make sure the following equation is correct, it looks a little strange...},
\begin{equation}
    \label{eq:intent_plan}
    % \hat{g}_{t+1} =arg\ max P(g_0,...,g_t|S_{seq})
    P(\hat{g}_{j} \vert \{g_{j-N}, ...,g_{j-1}\})
\end{equation}
where $g_i$ donates the $i-th$ requirement which has been confirmed and $\hat{g}_{j}$ donates the next potential requirement for conversation. 
% For example in Fig. \ref{fig:intent_space}, the user tries to express a series of intentions such as travel, airfare, restaurant, etc. 
The order of requirement expression is not unique. We assume that users with different PUS have different personalized habits to express all their requirements in unique order. These habits can provide dialogue system with effective tendency to plan the order of requirement confirmation. Most of requirement relationships are uncertain. Based on the $R\_Tree$ model, there are dependencies between requirements, such as restaurant--Peking Duck, etc. The dependency is presented in $R\_Tree$ that they are not in a topological order but a kind of parent-child node structure.

\subsubsection{Requirement Attribute Preference}
% We extract the user requirement attribute (RA) preference from dialogue corpus. 
Like the user profile~\cite{Minliu} 
% \my{have you mentioned user profile before?}
, RA module is the process of understanding users by extracting their interests on different constraints. Based on RA preference, the system can guide users to express their requirements in a more directional way during dialogue.
% 用户画像是基于用户属性和用户行为提取用户的兴趣，了解用户的过程。对 dialogue system 来讲，在对话时通过提 问了解用户的潜在需求，用户对不同服务和服务属性的拒绝和接受反映了其偏好。通过多次对话，dialogue system 不断加深对用户的理解，可以构建用户偏好的“用户画像”，用户与 dialogue system 交互过程中产生的对话是包含关于隐式和显式用户需求以及是否接受推荐服务的决定 的丰富信息的对话，换句话说，这些对话包含丰富的用户偏好。在和用户交互过程中， 基于此画像偏好能够更有方向性的引导用户表达意图。
% The restriction preference includes four factors: restriction type, restriction interval rank, restriction interval range, patience degree. \textbf{Restriction type} describes the importance users place on different restriction categories.\textbf{Restriction interval rank} describes the user's preferred interval rank to the restriction with interval category. \textbf{Restriction interval range} express the user's preferred interval range. And \textbf{the patience degree} donates the average rounds of dialogue to confirm an intention.
% 类似于用户画像，我们从语料库中抽取用户关于不同意图的约束类型的偏好。约束偏好包括四个方面{约束种类的偏好，约束区间等级的偏好，约束区间范围的偏好，对意图的耐心程度}。约束种类偏好描述了用户对不同约束类别的倾向程度。约束区间等级偏好表述了用户对于区间类别的约束，倾向的区间等级。约束区间范围偏好表述了用户对于区间类别的约束，倾向的约束区间大小。而用户耐心度表达了用户确认一个意图的平均对话轮数。
We design a key-value structure to describe the RA preference. Specially we use text mining method~\cite{Minliu}
% \my{too ambiguous... references or clarify what methods..}
to extract constraints of different requirements
% \my{what is useful information?}
from the dialogue corpus to generate or update the preferences. 
%% Mingyi: 这句的前后逻辑是在哪..我没太明白
% \my{see comment}
% The main idea is using user's response for different intentions to analyze their characteristics.
% 对于用户风格中的实体属性偏好，我们设计了基于键值结构的偏好风格模型，并使 用文本挖掘方法从对话语料中抽取偏好信息以生成/更新该偏好风格模型。偏好风格模 型的核心是“标注”，即利用用户行为信息来分析用户特征。

%% Mingyi: **只是建议**，我发现你在方法里面进行符号定义的时候总是从举玩具例子的角度进行，但是这个我个人觉得是不合适的（在introduction的时候合适），在具体解决方法的时候是否还是更多的从形式化的角度去阐述。。
% =======
% Demin: 下意识的想要说的清晰一些。类似的问题已调整删除
The RA preference can be expressed as $E = (U, P)$, where $U$ describes the general information of the user.
% such as gender, age, etc. 
$P$ represents the set of preferences for requirements attributes under different domains. The each identified requirement attribute preference $p_i$ can be expressed as follows:
\begin{equation}
    \label{eq:attri_preference}
    p_i = (Req, Name, Value, Freq)
\end{equation}
% 可以表示为 E = {U, P}。其中 U 描述了该用户的静态 信息，如用户的性别、年龄等。P 代表了用户对于不同领域下的实体属性的偏好集合。 针对每个确认过的实体属性的偏好 p，
%where $Intent$ represents the the domain of intention
% \my{what is the difference with the intent in Eq.1, if no difference, you dont need this sentence here.},
%such as the attraction domain, hotel domain, etc. 
$Name$ and $Value$ respectively donates the constraint name and value of preference of $Req$.
% represents the attribute value of constraint
%, such as rating level, price range, etc. 
%$Value$ represents the user preference for the current attribute. 
Here we divide the preference value into two types: Interval Value and Discrete Value. 
% Discrete value corresponds to intention attribute that can be presented as a discrete value, e.g., the hotel type is Deluxe, Normal. While the Interval Value can be presented in the form of an interval value, e.g., the price of \$100-\$200.
$Freq$ represents the frequency of requirement constraint appeared in the last limited rounds of corpus. It can reflect the importance of user to different constraint attributes.
% Name 表 示当前实体的属性值，如评分等级、价格区间等。Value 代表用户对于当前属性的偏好 期望值，这里我们将偏好值分为两种类型：区间型 Value 和离散型 Value。区间型 Value 对应的实体属性往往在对话中以区间形式被提出，如价格为 100-200 元；离散型 Value 对应的实体属性往往在对话中以离散值的形式被提出，如酒店类型为豪华型、普通型。 Freq 代表该实体属性在该用户最近有限轮语料中出现的次数，偏好被提及的频率能够反 映出用户对该类属性的重视程度。

%% Mingyi: 你这个framework和methodology
%======
%% Demin: Framework重点是介绍相关问题与模型定义，methodology则是针对模型提出具体方案
\begin{table*}

\centering
\caption{Updating Strategy of $Freq$}
\begin{tabular}{lll} 
\toprule
\multicolumn{1}{c}{Constraint scenarios} & \multicolumn{1}{c}{Strategy to update $Freq$} & \multicolumn{1}{c}{Example}                                                                                  \\ 
\hline
User initiated                                             & $Freq$ = $Freq$ + 2                         & User: I want to find a cheap hotel.                                                                          \\
User Accept                         & $Freq$ = $Freq$ + 1                         & \begin{tabular}[c]{@{}l@{}}Bot: Do you need a 5a  attraction?\\User:~ That's great!\end{tabular}  \\
User reject                         & $Freq$ = $Freq$ -1                          & \begin{tabular}[c]{@{}l@{}}Bot: Do you need a cheap hotel?\\User: No, I prefer a luxury one.\end{tabular}    \\
\bottomrule
\end{tabular}

\label{table:freq}
\end{table*}

\section{Methodology}
\label{sec:method}
\subsection{Personalized Utterance Style Mining}
\subsubsection{Act State Transfer Preference}

% 根据上文提出的对话行为体系，用户和dialogue system的所有对话都被标记对话行为，因此，围绕一棵意图树的对话的所有语句构成了一个对话行为序列 $Seq_{DA}=\{Act_{usr}^1,Act_{sys}^1,...,Act_{usr}^N,Act_{sys}^N\}$. 我们认为对话行为序列$Seq_{DA}$包含了用户潜在的对话动作倾向。例如，现实中不同用户对于自己的主观需求也可能存在不确切。当一个用户在表达寻找酒店需求后（State_in动作），会倾向于等待系统的建议(Resp动作)，而另外的用户则更倾向于直接表达自己对于酒店的约束条件（State_in动作）。
According to the $DA$ categories proposed in Table \ref{table:act}, all utterances are tagged with dialogue acts so that all $DA$s in a dialogue about specific $R\_Tree$ constitute a sequence of dialogue acts:
\begin{equation}
    Seq_{DA}=\{Act_{usr}^1,Act_{sys}^1,... ,Act_{usr}^N,Act_{sys}^N\}
\end{equation}
We consider that the $Seq_{DA}$ contains the potential act state transfer (AST) tendencies. 
Users with different $DA$ preference tend to choose different response $DA$ when they in the same state of context.
% \my{too much for example....}
% For example, different users may also have inexact subjective requirements for themselves. While one user, after expressing the requirement to find a hotel ($State\_in$ action), would tend to wait for the system's suggestion ($Resp$ action). While another user would prefer to directly express his constraints for the hotel ($State\_in$ action).
In this work, we use weighted finite state transducers (wFST)~\cite{zhou_augmenting_2019} to capture the latent AST preference. Unlike RNN, wFST can explicitly track the entire path it traversed, which gives additional symbolic constraints and information about the dialog history. A trained 
% \my{trained or pre-trained?}
wFST serves as a scaffold for dialog history tracking. Meanwhile, wFST is more interpretable, as each state is explicitly represented by an action distribution which is easier for humans to interpret model decision. As part of PUS, the AST module can model the dialog acts states to generate response act according to request. 
Given all $DA$ sequences of one user, we use the greedily node splitting algorithm\cite{zhou_augmenting_2019} to train the initial wFST
%\my{but you say you use pre-trained wFST, why here you train the initial wFST, what you actually do?}
. For a trained wFST, we can understand the dialog state node meaning by checking input and output edges. Given the current dialogue act sequence $\{Act_{usr}^1,Act_{sys}^1,...,Act_{usr}^{t-1}\}$,  the last user dialogue act $Act_{usr}^{t-1}$ is fed to wFST, and then it gives a probability density for the likelihood of transfer from the current state to each of the possible dialog acts. Compared to RNN, the wFST can not only return the current state embedding but also all information of the state it traversed since the start state. 

\subsubsection{Requirement Planning Preference}
Users with different PUS preference differ in the requirement expression order even within the same dialogue goal. % \my{for example again....}
% For example, user may prioritize all intentions around a specific domain, or they may choose to identify high-level intentions for all domains. 
Intuitively, given an $R\_Tree$ as the initial user requirements, the process of user requirement expression can be regarded as the node travelling for all requirement nodes of $R\_Tree$, and there are different expression orders of requirement sequence. 
% \my{????}
% the process of requirement expression may be in the order of depth priority or breadth priority, etc.
% All these orders belong to the personalized preference in intention expression. The dialogue system has sufficient knowledge
% \my{why can have sufficient knowledge, an assumption? or a common sense?}
%of the user's intention preference, so that it can actively and naturally guide the dialogue to recommend potential intentions to meet the user requirements. 
Given the sequence of requirements in the current context, dialogue system should continuously plan the next requirement to be identified, and the candidate requirements are chosen based on the potential patterns of requirement sequence in dialogue history. Therefore, we consider requirement planning as a simplified Sequence Recommendation (SeqRec) task~\cite{rendle2010factorizing}.
% \my{paper cite}.
% 如前文所提到的，面对同一个对话目标，针对里面所有的意图点，具备不同对话风格的用户，在表达意图次序方面存在不同。例如，用户可能优先确认围绕某个领域的所有意图，也可能首先确认所有领域的高层次意图。直观的看，对于树形结构的对话目标，在对话过程中表达意图次序是，可能按照深度优先顺序或者广度优先顺序，甚至跳跃式表达意图，这些都属于用户在表达意图时的个性化倾向，这里的每个意图点都代表的对话话题的转移。dialogue system对用户意图偏好有足够认知，便可以主动且自然的引导对话方向进行候选意图推荐，从而满足用户需求。给定当前对话上下文的对话意图序列，dialogue system需要持续地规划下一个待确认的意图点，候选意图点选择的依据是根据该用户所有意图表达序列中的潜在模式。因此，我们将该任务视为简化的序列推荐(SeqRec)任务.

The classical SeqRec tries to model the sequential user behaviors with items, mining the connection between user contexts, requirement, and goals for more accurate and customized recommendations. 
As a simplified version of the sequence recommendation task, we only need to focus on the recommendation of requirement sequences for different users, so we use Factorizing Personalized Markov Chains (FPMC)\cite{rendle2010factorizing} to learn the latent preferences between users and requirement sequences. FPMC is designed by personalized transfer on incomplete Markov chains, which combines the features of both matrix decomposition and Markov chain recommendation models. Specifically, it combines the personalized set Markov Chains with the factorized transition cube results
% as the recommender
% \my{typo: recommender}
to capture user preferences over requirement-to-requirement transitions.
Given all requirement sequences of all users, we use the S-BPR algorithm\cite{rendle2010factorizing} to train a personalized transition cube to learn the requirement expression preference of different users. For the requirement planning task, FPMC can effectively solve the problems of user-requirement data coefficients and serialized local dependency modeling.
% 给定所有用户的所有的对话意图序列，我们使用S-BPR算法训练Personalized transition cube，从而学习不同用户的意图表达风格。针对对话序列规划任务，FPMC可以有效解决用户-意图数据系数以及序列化局部依赖性建模的问题。

% \begin{figure*}[htp]
% \centering
% \subfigure[\scriptsize{Baseline}]{
% \begin{minipage}[htp]{0.9\columnwidth}
% 	\includegraphics[width=1.0\columnwidth]{figures/Framework_baseline.png}
% 	\label{fig:framework_baseline}
% \end{minipage}
% }
% \centering
% \subfigure[\scriptsize{Ours}]{
% \begin{minipage}[htp]{0.9\columnwidth}
% 	\includegraphics[width=1.0\columnwidth]{figures/Framework_ours.png}
% 	\label{fig:framework_ours}
% \end{minipage}
% }
% \caption{Framework of Dialogue System}
% \label{figure:framework}
% \end{figure*}

\subsubsection{Requirement Attribute Preference}

We mine requirement attribute preference from the user dialogue corpus. The essential task of preference mining is to effectively count the $Freq$, which represents the frequency of the attribute that appears in the recent limited rounds of the user's corpus. Frequency can reflect the importance that users place on this constraint. For user's last $N$ rounds of dialogue, the $Freq$ domain of each preferred attribute is continuously updated as the conversation progresses, which also reflects that the user's attribute preferences are keeping changing. 
% Dialogue system focuses on the user's attribute preferences at the current moment.
For each user's dialogue corpus, we analyze utterance of each round. For each utterance, the triples of dialog action, defined at equation \ref{eq:understand}, can be extracted with natural language understanding (NLU) model or manual rules.
%using the intention understanding model to obtain the intention structure\my{what is intention structure??? metioned-above?}. 
Since the scenarios in which entity attributes appear can be either active expressions of the user or assisted system recommendations, different scenarios for the appearance of entity attributes need to update $Freq$ according to different strategies. We design the $Freq$ update strategy as shown in Table \ref{table:freq}.
% \my{table should be explained.}.
Users' preferences on requirement attributes can be reflected by their rejection or acceptance to different constraints.
The $Freq$ would upgrade when the constraint is approved. And the $Freq$ would be decreased if user reject the constraint.

% 我们从用户对话语料中挖掘实体属性偏好。偏好挖掘的最主要工作是有效统计Freq，它代表该实体属性在该用户最近有限轮语料中出现的次数，被提及的频率能够反映出用户对该类属性的重视程度。对于每个用户最近N轮对话的数据，每个偏好属性的Freq域是随着对话的进行不断更新的，这也反映出用户的属性偏好是会随着时间而不断变化的，而dialogue system关注在当前时刻下的用户属性偏好。
%对于每个用户的对话语料，我们利用意图理解模型分析每一回合的人机对话文本，获得用户和系统的意图结构。由于实体属性出现的场景可以是用户的主动表达，也可以是系统的辅助推荐，因此对实体属性出现的不同场景需要按不同的策略更新Freq。对于Freq更新方法我们设计了如下策略：
% \usepackage{booktabs}

\begin{figure*}[ht]
\centering
\includegraphics[height=1\columnwidth]{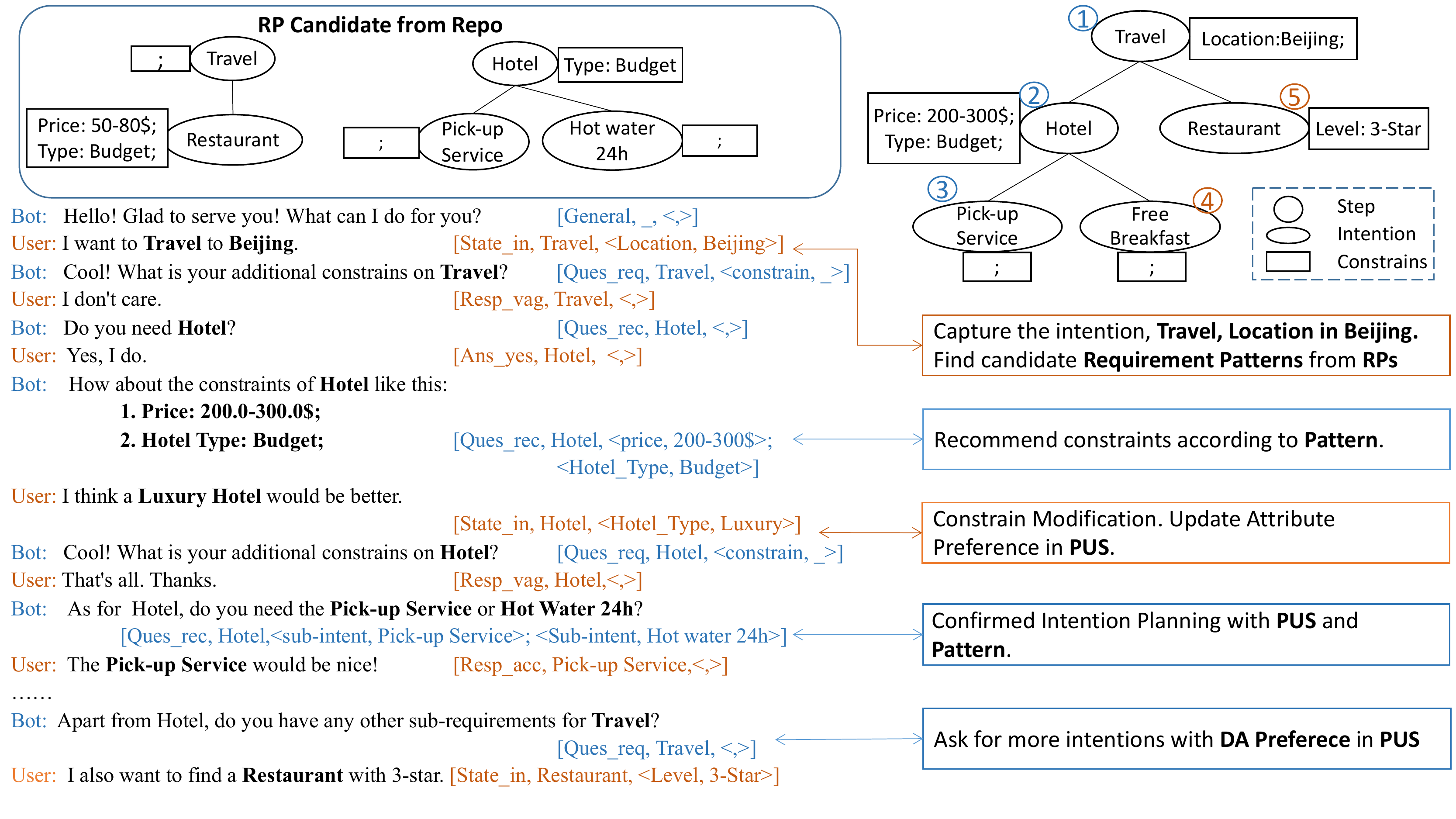}
\caption{The Example of Dialogue Strategy}
\label{fig:dialog_example}
\end{figure*}

\subsection{Multi-Round dialogue for Requirement Elicitation}

\subsubsection{Requirement State Updating based on Requirement Pattern}
% 在现实世界中，用户的很多需求可以不分解，因为这些需求已经有了响应的解决方案。

% The goal of the dialogue system in this work is to elicit the user's complete intentions with multiple dialogues. 
% 本文中对话系统的目标是通过对话分析用户的意图，得到用户的目标意图树。
In traditional task-oriented dialogue scenarios, dialogue state is usually updated by static slots, which is the basis for dialogue goals.
% 在普遍的面向任务的简单对话场景中，通常对识别的用户意图使用插槽的方式更新对话状态，往往通过ontology-based方法补充对话状态中的缺失部分，最终根据固定的槽位信息提供服务。
These strategy will get impressive results in specific dialogue scenarios, such as fight enquiry. However, because of the complex requirements in SRE task, the strategy based on static-slot
%\my{fixed-slot or static slot?}
filling cannot play to their advantage.
% 这种策略面向固定场景的特定任务能够获得很好的效果，但面对比较特殊的需求获取任务，由于复杂的用户意图往往会涉及到不同的领域场景，并且需求的结构也不尽相同，因此这种特定槽位的对话策略不能发挥他们的优势。
As for SRE task ,the $R\_Tree$ can effectively modeling complex requirements in multiple domains and won't be limited by fixed-slot in specific scenarios, which provides a foundation for dynamically requirement building. 
% 针对需求获取任务，意图树模型能够有效对用户的多领域复杂需求建模，其形式不会受到领域信息限制，为动态构建用户意图提供了基础。
%There is an example of requirements tree shown in Fig. \ref{fig:dialog_example}. 
% we use the $R\_Tree$ to build the user's complete intentions，which is a necessary part for subsequent service solution design. 
% 为了适应不同领域的需求获取任务，我们利用不同领域的需求模式辅助对话策略，为需求引导动态地提供的候选槽位。

In the real world, there will be lots of redundant
% \my{recurring? do you mean repeating or redundant}
requirement fragments appearing together on different $R\_Tree$ in similar environment. 
% 然而在现实生活中，对于相似外部环境下的多个用户意图树，往往会存在频繁出现的重复的需求片段。以城市旅行意图为例，大部分游客的意图往往会将餐馆与旅行意图相关联，
% \my{example again..}
As an example about travel in a food city shown in Fig. \ref{fig:dialog_example}, most requirements of tourists will associate famous restaurant with the node of travel. 
These requirement fragments frequently appear together in user’s requirement trees are called requirement patterns ($RP$) \cite{tian2022intention}, which can be aggregated to form new coarse-grained requirements.
% 这些在用户意图树中频繁出现在一起的需求片段被称为需求模式，能够被聚合成聚合到一起形成新的粗粒度的意图。
These requirement patterns exist in different fields and can be used as the basic components to provide support for requirements mining.
% 这些需求模式存在于不同的领域，能够作为需求的基本组成部分，为需求挖掘提供支撑。
As for service provider, these requirement patterns do not need to be decomposed because there is already specific solution for those requirements~\cite{xu2020domain}.
% \my{paper cite}
% 对于服务提供者，这些需求模式可以不必被分解因为已经存在对这些需求的的解决方案。

% 在本文中，我们使用意图树来构建用户的整体需求，一棵完整的意图树的示例如图4所示。在获取用户的完整需求后，下游的服务提供者由此配置合适的解决方案提供给用户。
Base on $RP$, we can effectively address the domain-limited problem of dialogue state in dialogue and realize the dynamic expansion of the dialogue state.
% 利用需求模式，我们可以有效的解决对话过程中对话状态的领域受限问题，实现对话状态的动态拓展。
% 我们在Tian et al 工作的基础上，设计了基于需求模式匹配方法的动态需求获取对话策略。我们认为对用相同领域的意图结果
We use requirement pattern mining algorithm\cite{tian2022intention} to extract frequent requirement patterns from requirement trees in specific domain, and apply a RP-based requirement steering strategy to capture user requirements.
% 我们使用基于需求模式辅助的需求引导策略来获得用户意图。
Specifically, we store all $RP$ from different domains in requirement pattern library ($RPs$) to provide necessary domain information. In dialogue process, once the basic requirement is captured, the $RP$ in pattern library is matched according to requirement. Then we use static information in user portrait, such as external environment, time and place, and user attribute preferences to choose candidate $RP$ that meet user requirements. Using $RP$, system can choose different dialogue actions to perceive the possible expansion direction of current requirement and plan the direction of conversation topic. Finally, requirement reconstruction is completed with assistance of $RP$.
% 具体来说，我们将所有来自不同领域的需求模式存储于需求模式模式库用来提供必要的领域的先验的信息。在对话过程中，一旦捕捉到用户的基本意图后，根据该意图从需求模式库中的需求模式作匹配，并依赖用户画像中的静态信息，如外界环境、时间地点等以及用户属性偏好，筛选出符合用户需求的候选需求模式。利用需求模式，系统感知到当前上下文中可能的意图拓展方向，来规划对话话题的走向，引导用户表述其真实意图。最终以需求模式作为对话依据和用户交互来完成需求重构的完整过程。
$RP$ can be extracted from multiple fields in an offline state, and be used for real-time requirements adaptation using user profile. The dialogue topic guidance strategy based on $RP$ can effectively realize adaptation process between multiple dialogue fields. As for strategy based on fixed slot filling, users can only choose to offer their requirements directly when there is no domain slot information. As the complexity of requirement increases, the unlimited fixed slot filling strategy will lead to rapid growth of dialogue rounds, leading to terrible dialogue experience.

\subsubsection{Incorporating PUS into Requirement Elicitation Strategy}

\begin{figure}[ht]
% \centering
\includegraphics[width=\linewidth]{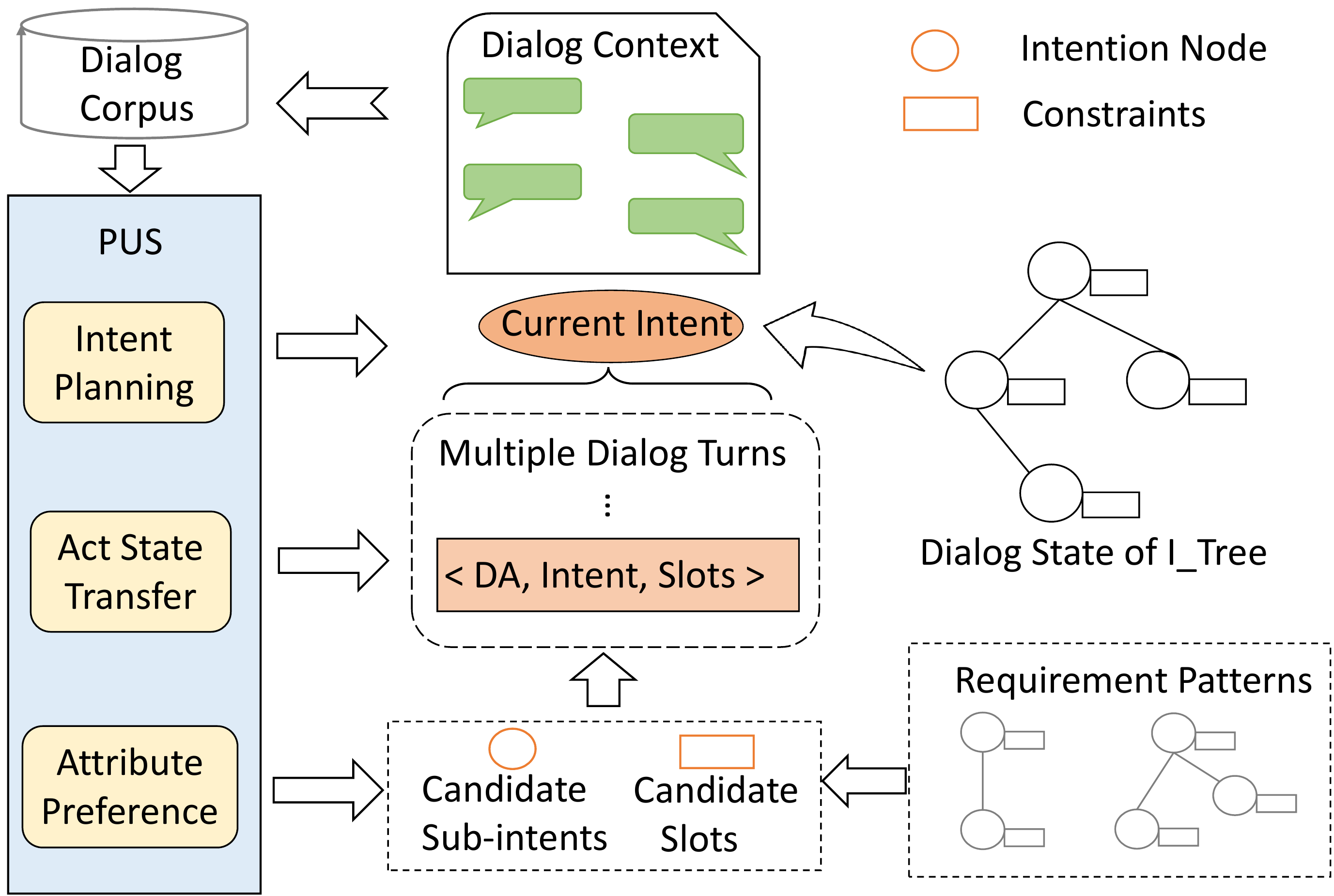}
\caption{Dialogue Strategy with PUS}
\label{fig:dialog_pus}
\end{figure}

%，其完成的功能可以概述为利用有频繁挖掘获得的需求模式构成的需求模式库，补充完善当前需求树模型使之能够描述完整需求。接下来我们需要将提出的用个性话对话风格模型结合需求模式，作用在对话策略中以完成需求获取任务。
%面向需求获取的对话系统，其结构与通用的任务型对话系统的结构类似，除了第2.2小节中提到的用户意图理解模型，还包括对话状态更新模块、对话策略模块以及自然语言生成模块。融合个性化对话风格的需求获取方法的主要目的在于针对用户的请求决定系统下一步的响应方式，这便是对话管理模块的主要功能。对话管理模块在确定系统相应行为时，需要综合考虑用户状态、用户情绪以及对话过程的通用行为模式，继而选择最佳反馈行为以实现高效的用户需求引导。
%融合对话风格的需求获取方法过程如算法1所示。对于整段对话，我们将对话过程根据对话意图与意图约束两个层面抽象为内-外双循环结构。对话的整个过程在于首先确认某项话题（意图点），然后围绕该话题展开不同细节讨论。外层循环即负责确认并规划对话话题，利用用户的对话风格中对于意图点讨论的偏好顺序以及需求模式匹配方法，规划整段对话的话题走向，这种意图规划的方式是基于意图树结构中的节点遍历方式而设计。而内循环则针对某个话题，以回合对话的结构确定每个对话回合中，用户的意图与系统反馈。在内层循环中会随时根据系统当前状态与用户意图，判断对话话题是否发生改变，进而实现内外循环间的转移。算法基于对话意图公式公式2-1描述对话意图结构。在对话过程正，系统维护意图树I_Tree_{cur}作为系统对话状态，真对用户的每一次请求的对话意图，会更新当前意图树的结构，而结合当前对话状态与用户对话风格，系统实现对于响应动作的预测。Predict\left(I_Tree_{cur},PUS,Node_{candi}\right)函数的主要功能即针对当前系统意图树状态，结合用户对话风格，这里主要分析用户对话动作偏好，生成系统的最优响应动作。在这之前，对于待确认的属性约束选择上，系统会充分考虑需求模式匹配的候选约束以及用户对于不同意图的属性偏好综合权衡。最终当意图规划模块不再存在候选意图或者用户提出主动结束对话，对话过程终止。此时系统维护的对话状态意图树即作为系统对于用户意图的全面理解。
We design a dialogue strategy based on PUS to elicit user requirements more efficiently.
The main target of PUS incorporation can be divided into two parts. Firstly, determine the next dialogue act response to the user request. Secondly, plan direction of dialogue topics with personalized preference. The overview of dialogue strategy with PUS is shown in Fig. \ref{fig:dialog_pus}. 
% 基于需求模式匹配的对话策略，pus的主要目标可以被分为两部分：1）根据用户请求确定下一步的系统相应，这是对话管理模块的主要功能。2）利用个性化偏好规划对话话题的方向

For the entire requirement elicitation process of dialogue, we abstract the dialogue process into an inner-outer loop process according to requirement of $R\_Tree$ and  requirement constraints. 
Based on the Requirement Planning in PUS and $RP$ matching, the dialogue strategy plans the order of requirement confirmation in the outer loop. Throughout the whole dialogue process, every requirement in $R\_Tree$ would have different states for the dialogue system: potential requirement in space, candidate requirement for planning, requirement accepted, current requirement in context, requirement confirmed. 
All requirements before dialogue can be regarded as a huge potential requirement space, shown in Fig. \ref{fig:intent_space}. Then based on $RP$ matching, the system would search some requirements as candidate requirements for requirement guidance. When user proactively proposes requirements or accepts the requirements from system recommendations, the requirement of requirements can be accepted. According to the requirement in the current context, there will be several rounds to discuss constraints about current requirement. When all constraints about current requirement have been confirmed, the system will choose another topic to continue the dialogue.

\begin{algorithm}
% \small
    \textsl{}\setstretch{1.2}
    \caption{Algorithm of Requirement Elicitation Strategy}
    \label{alg:dialog_pus}
    \begin{algorithmic}[1] % 显示行号
        \Require $RPs$ % ;$R\_Tree_{s}$;$PUS(user\_id)$
        \Ensure $R\_Tree_{t}$
        \State $Reqs_{c} \gets$ Initial $rp$ requirement extract from $RPs$
        \For{ $len(Reqs_{c}) > 0$}
            \State $Re\_Req \gets PUS.RP(R\_Tree_{c}, RPs)$
            \For{$not\  Topic\_Change(Context)$}
                \State $<DA, Req_{cur}, Slots> \gets NLU()$
                \State Update $R\_Tree_{c}$ State
                \State $ Re\_DA \gets PUS.AST(DA,R\_Tree_{s})$
                \State $Re\_Slots \gets PUS.RA(RPs, Slots)$
                % \STATE Update $Reqs_{c}$ with Response Action.
                \State NLG($Re\_DA,Re\_Slots$)
            \EndFor
            \State $Reqs_{c} \gets Update(RPs, Re\_Req, R\_Tree_{s})$  
        \EndFor
        \State $R\_Tree_{t} \gets R\_Tree_{s}$ \\
        \Return $R\_Tree_{t}$
    \end{algorithmic}
\end{algorithm}

As for each round of dialogue, the dialogue system needs to determine the response action, described in equation \ref{eq:understand}, according to user state and dialogue context. We use the personalized act state transfer preference to predict the $DA$ of response action. User profiles and constraints in $RP$ can be equipped to generate suitable $Slots$ of response. In the inner loop, the strategy will also judge whether the current requirement has changed according to user action and system state of $R\_Tree$, so as to realize the transfer between inner and outer loops. 
The algorithm \ref{alg:dialog_pus} describes the whole dialogue strategy for SRE.
The $Predict\left(\right)$ function is to focus on the current system state and PUS to analyze user action preference, and generate optimal response action. The dialogue process is terminated when the requirement planning module no longer has candidate requirements or the user proposes to end the conversation. The dialogue example shown in Fig. \ref{fig:dialog_example} demonstrates the process of requirement elicitation of our dialogue strategy.

\section{Experiments}

\subsection{Experimental Setup}
\subsubsection{SRE Task Setup}
The SRE task is oriented toward scenarios where dialogue system interacts with user and guides them to express complete requirements. For each conversation, user owns their requirements as dialogue goal, which can be regarded as a $R\_Tree$. Then users express requirements or constraints of every requirement node through multiple dialog rounds in a personalized manner. Users can express their requirements either actively or passively. Then the dialogue system captures user's requirements and updates the target $R\_Tree$. 
The conversation stops when all user requirements have been confirmed or the dialog round reaches the max number, which is set as 17 for the average dialogue corpus.
Otherwise, the conversation is regarded as failed.

We re-implement the requirement-oriented dialogue strategy from Tian et al.\cite{tian2022intention} , which also applies requirement tree to dialogue state and incorporates external knowledge into requirement discovery.
To achieve personalized interaction with dialogue system, we use an Agenda-based user simulator \cite{schatzmann-etal-2007-agenda} to simulate user behaviors. 
It uses a compact representation of the user goal of $R\_Tree$ to guide dialog topic. 
Given a dialogue example with requirement tree, the simulator analyzes $R\_Tree$ as dialogue goal. Then, it pushes different requirements and constraints into a stack structure in the order of their appearance in the original dialogue.
% 给定数据集中的一段对话，用户模拟器解析意图树作为对话目标，并将对话过程中不同意图、约束按照在对话中出现的先后次序压入栈结构中
At each turn, the simulator receives system dialogue actions, modifies its state, and outputs user dialogue actions according to specific hand-crafted rules.
Both our pus-based dialogue strategy and requirement-based dialog strategy \cite{tian2022intention} interact with simulator to evaluate dialogue effectiveness.
% 需求获取任务面向的是dialogue system与用户对话交互，引导用户表达完整需求的场景。对于每一段对话，用户被提供本次交互的对话目标，即初始意图树。用户按照个性化的方式，通过多轮对话向dialogue system主动或被动的表达意图树中的意图和约束。dialogue system通过与用户的交互感知用户意图并构造目标意图树。当系统检测到用户意图已完全实现后便可以终止此次对话。
\subsubsection{Dataset Annotation}
% \subsubsection{Dataset re-organization}
We use CrossWOZ\footnote{https://github.com/thu-coai/CrossWOZ}, a task-oriented conversation dataset, to extract the utterance style of different users.%The corpus is to simulate scenarios where a traveler seeks tourism information and plans her or his travel in Beijing.
 As for SRE task, dialogue system no longer does entity-oriented search actions, but focuses on understanding and guiding the user requirement expression. Therefore, we modify and delete question-answer pairs in original dataset that involve entity search with manual rules.
% For the above example, the modified dialogue goal will become: the user's travel needs a five-star attraction and a nearby hotel, and travels by cab. For bot, the conversation ends with an intent tree for this requirements. From the perspective of a complete dialogue system as service provider, the entity recommendations for the user's requirements will be executed by the subsequent recommendation module.
% 对于我们的需求获取任务，对话系统不再去做面向实体的搜索动作，而是关注与对用户意图的理解与引导表达。因此我们基于一定规则和人工的方式，将原始数据集中涉及到实体搜索的问答对作修改或删除，改造成需求获取任务。针对上述实例，修改后的需求获取任务将会变成：用户的旅行需要五星级的景点以及一个临近的酒店，并以搭乘出租车的方式出行。对bot而言对话结束后会生成针对这一用户需求的意图树。从完整的dialogue system系统角度，针对用户需求的实体推荐会由后续推荐模块专门执行，本文的对话系统只关注第一阶段的需求获取任务。
 Although the original dataset was not designed with personalized characteristics, the workers who played the USER role were encouraged to dialogue on their terms when collecting the data \cite{zhu2020crosswoz}.
%modify the relevant constraints according to their personal preferences in some cases
%, in addition to being asked to fill the relevant slots and generate dialogue requests according to the dialogue goals.
%According to the observed dataset, 
% Even given the same dialogue goal, different workers choose different orders and strategy to complete goals.
%when realizing the dialogue task, which is also consistent with the usual dialogue experience. 
It is important to emphasize that the PUS preferences mentioned in this paper mainly refer to the workers' habits when expressing their requirements, rather than just attribute preferences in other user-profile related studies. 
For the SRE scenario, we use the dialogue act categories shown in Table \ref{table:act} to re-annotate dialogue acts in CrossWOZ. We also use manual annotation to assist in labeling cases that cannot be handled by the rules if necessary. 
\subsubsection{Evaluation Metrics}
% 对话次数表示对话系统识别用户需求的过程中的交互频率。通常情况下，复杂和冗余的对话可能会导致糟糕的用户体验。换句话说，对话次数越少，用户体验越好。因此，本研究以 "对话圈数 "作为评价指标，即具有相同规模意图的对话的平均圈数。
The number of dialogue rounds indicates efficiency of interaction during process wherein dialog system identifies user requirements. Typically, complex and redundant dialogues may result in a bad user experience. In other words, the fewer the dialog rounds, the better the user interaction. Thus, this study takes \textbf{Dialogue Round}, the average rounds of dialogues with the same scale requirements, as an evaluation index.
% PRF指标一直被用来评价多轮对话策略的质量。在本研究中，我们使用PRF来评价多轮对话产生的意向树的结果。Recall指的是通过对话捕捉到的正确意图在用户真实意图中的比例。
% The Recall and F1-score indicator is widely used to evaluate the quality of multi-round dialogue strategy. 
In addition, we uses \textbf{Recall} and \textbf{F1} 
% \my{why not Precision, and why P in Eq.5}
to evaluate results of generated $R\_Tree$.
Considering that every requirement nodes in target $R\_Tree$ needs to be confirmed by user (requirement recommendation or statement), 
there would not be false requirement, and the Precision would always be 1.
The PRF of $R\_Tree$ can be calculated by following equations.
\begin{equation}
    P_{tree} = \frac{R\_Tree_{init} \cap R\_Tree_{gen}}{R\_Tree_{gen}}
\end{equation}
\begin{equation}
    R_{tree} = \frac{R\_Tree_{init} \cap R\_Tree_{gen}}{R\_Tree_{init}}
\end{equation}
\begin{equation}
    F1_{tree} = 2 \times \frac{P_{tree} \times R_{tree}}{P_{tree} + R_{tree}}
\end{equation}
Where $R\_Tree_{init}$ indicates the initial $R\_Tree$ of user requirements and $R\_Tree_{gen}$ indicates the $R\_Tree$ generated by dialogue system with conversation.
Finally, we use \textbf{Success Rate} to indicate the effectiveness of dialogue system. 
% 最后，我们使用对话成功率，来代表系统的性能。其中超过最大对话轮数而没有

\subsection{Main Results}

\subsubsection{Dynamic Requirement Elicitation with PUS }

\begin{figure}[htp]
\centering
\subfigure[The $R\_Tree$ accuracy of requirement-based policy.]{
\begin{minipage}[htp]{0.45\columnwidth}
	\includegraphics[width=1.1\columnwidth]{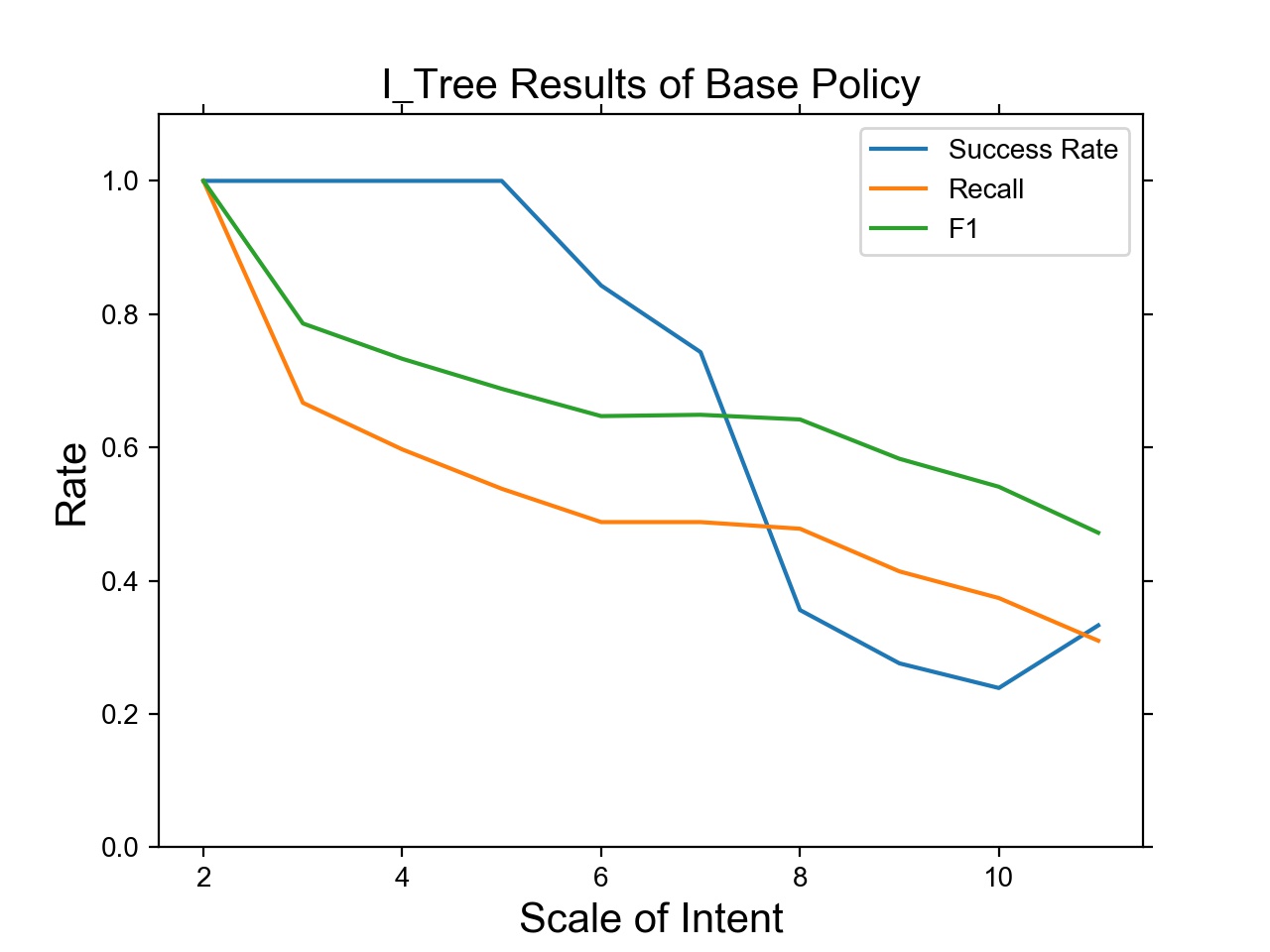}
	\label{figure:pus_strategy}
\end{minipage}
}
\centering
\subfigure[The $R\_Tree$ accuracy of PUS-based policy.]{
\begin{minipage}[htp]{0.45\columnwidth}
	\includegraphics[width=1.1\columnwidth]{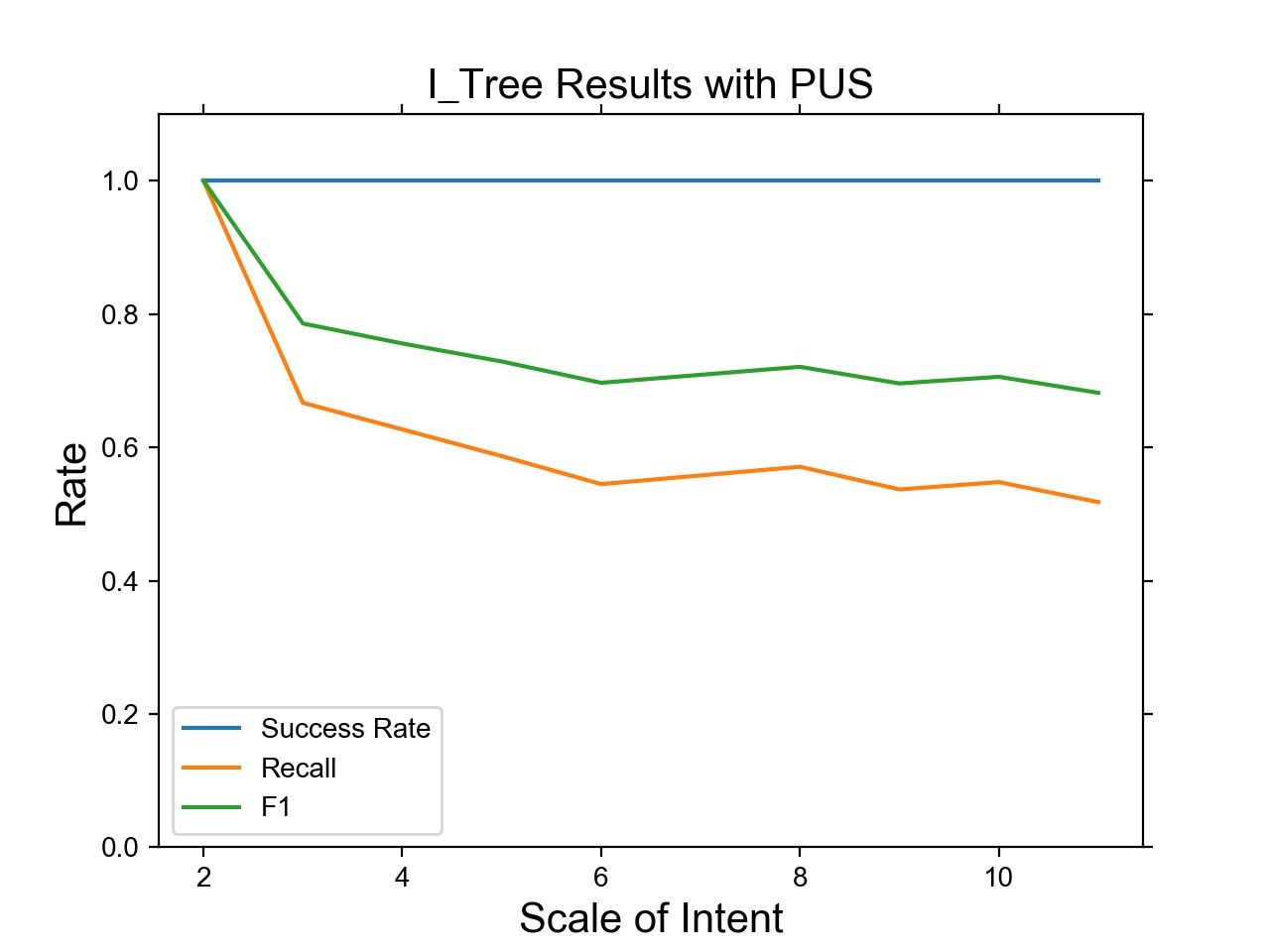}
	\label{figure:base_strategy}
\end{minipage}
}
\centering
\subfigure[Average dialogue rounds of two strategy.]{
\begin{minipage}[htp]{0.5\columnwidth}
	\includegraphics[width=1.1\columnwidth]{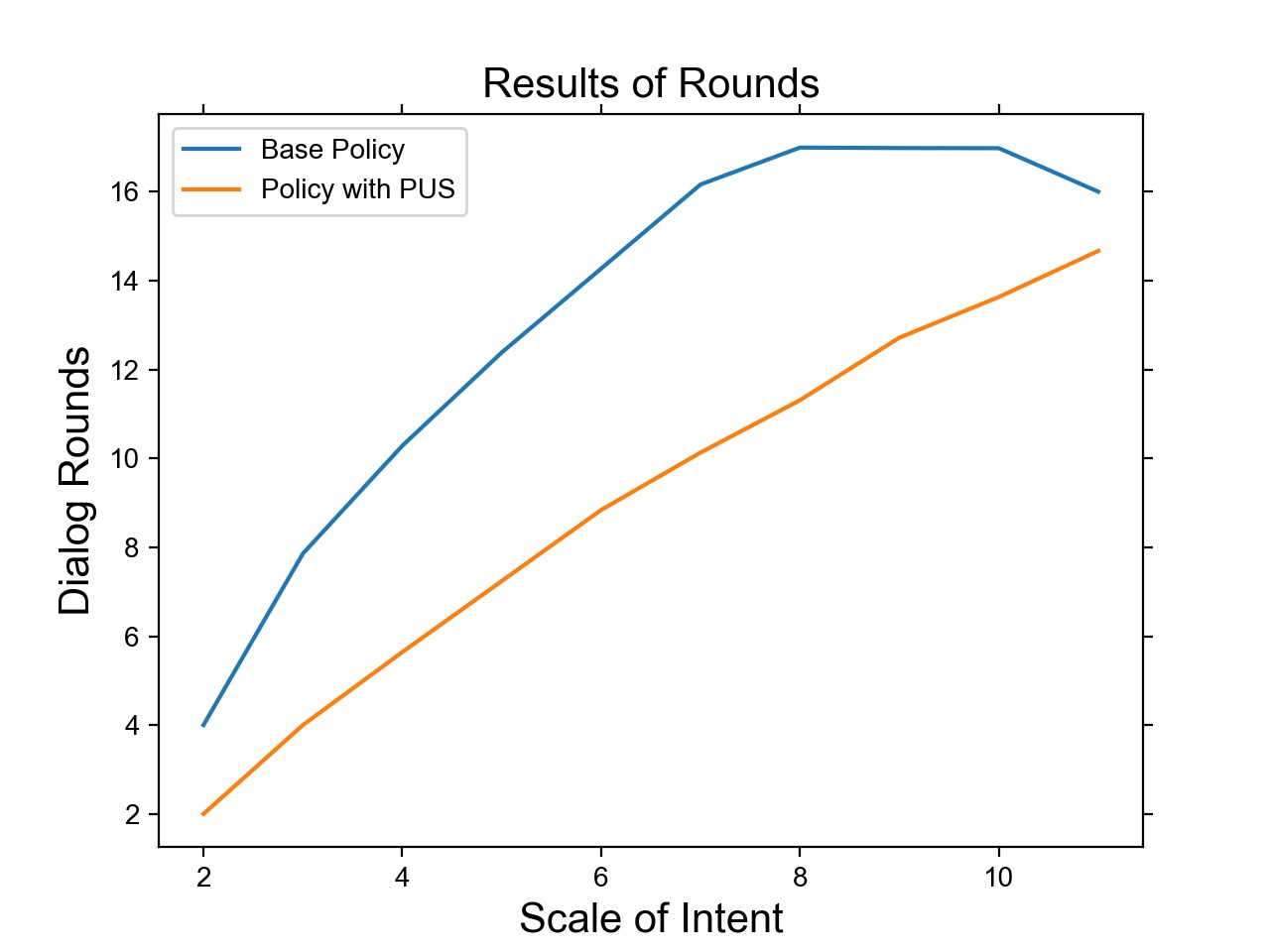}
	\label{figure:rounds}
\end{minipage}
}
% \my{I find the success rate of I_tree with PUS keeps to be 1, but you didn't explain it in your text,,, does this really fair??} 
\caption{The Results of Requirement Elicitation}
\label{figure:iTree_results}
\end{figure}

For the SRE task, we use the re-annotated CrossWOZ dataset to complete dialogue experiments. The user simulator uses current dialogue state and different personalized weights to simulate different dialogue styles. Different weights can achieve different action selection tendencies according to dialogue records. Dialogue system updates dialogue state and determines response action for each user action request, depending on user's PUS and $RP$. We apply our requirement-based dialogue strategy \cite{tian2022intention} and our pus-based dialogue strategy on dialogue 
simulation platform \cite{zhu2020convlab} and interact with dialog simulator for multi-turn dialogues.

After completing 4988 dialogue tests from 67 users, we obtained experiment results as shown in Fig. \ref{figure:iTree_results}. Fig. \ref{figure:pus_strategy} and Fig. \ref{figure:base_strategy} show accuracy of $R\_Tree$  and dialogue success rate under different requirement scales. It is worth noting that accuracy of dialogue requirements includes not only requirement nodes, but also constraints of different requirements. Fig. \ref{figure:rounds} shows average number of rounds per dialogue for different requirement scales.
% 对于需求获取任务，我们利用重新标注的CrossWOZ数据集中完成对话实验。我们设计了基于意图树作为对话目标的用户模拟器，根据当前的对话状态以及不同的个性化权重，这里的权重能够根据对话记录实现不同的动作选择倾向，以此来模拟不同的对话风格。这里我们依赖原始数据集中的用户id区分用户角色。只有具备足够对话语料的用户数据被用来作对话测试。对话系统根据用户的语料建模了对不同用户的pus认知。在对话过程中对话系统针对每次的用户动作请求，依赖pus和需求模式，更新对话状态并确定响应动作。当系统认为所有意图已经被确认，或者对话轮次超过最大轮次时，对话结束。我们设置最大对话轮次为17，当对话轮数超过17而系统并未完全完全意图，对话失败，因为过多的对话轮次容易获得用户糟糕的对话体验。我们将基于需求模式的对话策略作为baseline，与融合pus的对话策略作对比。在完成4988例对话后，我们获得了如图所示的实验结果。其中a,b展示了在不同意图规模下，系统获取的对话意图的准确程度以及对话成功率。值得注意的是对话意图准确程度不仅包括意图节点，还包括不同意图的约束是否与用户需求一致。c展示了不同意图规模下，每段对话的平均轮次。
The results show that the dialogue strategy incorporating pus can obtain more complete and accurate user requirements with fewer dialogue rounds, and can guarantee the success of dialogue under medium requirement scale. The main reason of difference is that system can take different dialog actions in specific context and search different directions in requirement space. According to requirement planning and action state selection in PUS, the dialogue system can select a more suitable direction with a higher probability. However, due to limitation of maximum number of rounds, the dialogue strategy that only relies on the $RP$ can only attempt different search states in limited steps. The dialogue success rate is relatively low when there are too many potential requirements.
Furthermore, since we set a large enough maximum number of dialog rounds, the dialog of strategy with PUS can confirm all requirements, indicating that the strategy can choose a more suitable response action with assistance of PUS.
% 结果表明，融合pus的对话策略能够用更少的对话轮次，获得更加完整准确的用户意图，而且在中等意图规模下能够保证绝对的对话成功率。这里的主要原因在于，对用户的需求引导类似于用户需求的搜索，在特定上下文中系统采取不同的对话动作面临不同的意图搜索方向。而根据pus中的意图规划和动作状态选择，系统能够更高概率的选择最优搜索方向。而只依赖需求模式的对话策略由于最大轮次的限制，在有限的搜索步骤中只能依赖需求模式试探不同的搜索状态，在意图过多时，对话成功率相对较低。

\subsubsection{Ablations of Personalized Features}

\begin{figure}[htp]
\centering
\subfigure[Response action with Global Tendency]{
\begin{minipage}[htp]{0.45\columnwidth}
	\includegraphics[width=1.0\columnwidth]{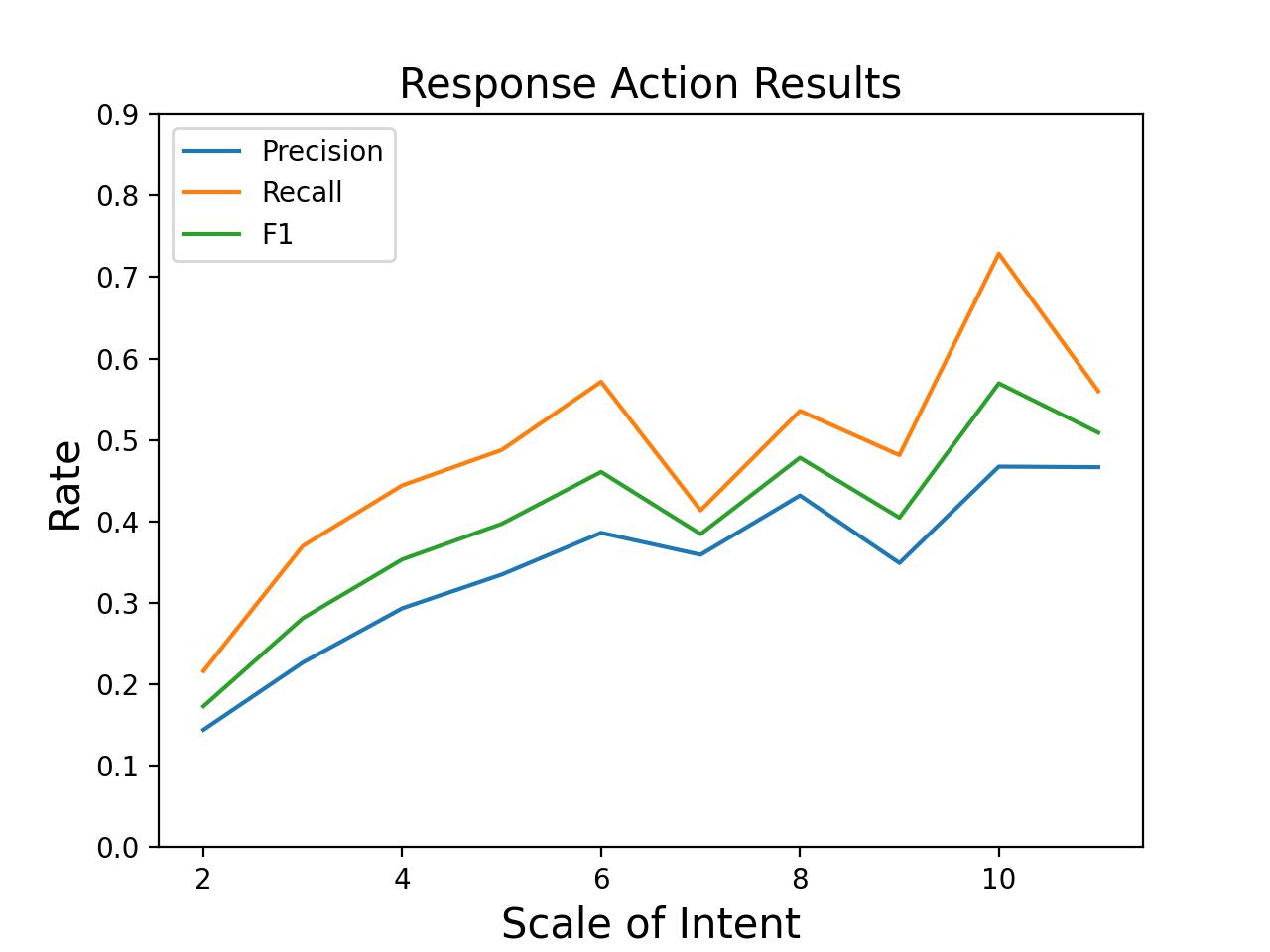}
	\label{figure:scale_patience}
\end{minipage}
}
\centering
\subfigure[Response action with PUS]{
\begin{minipage}[htp]{0.45\columnwidth}
	\includegraphics[width=1.0\columnwidth]{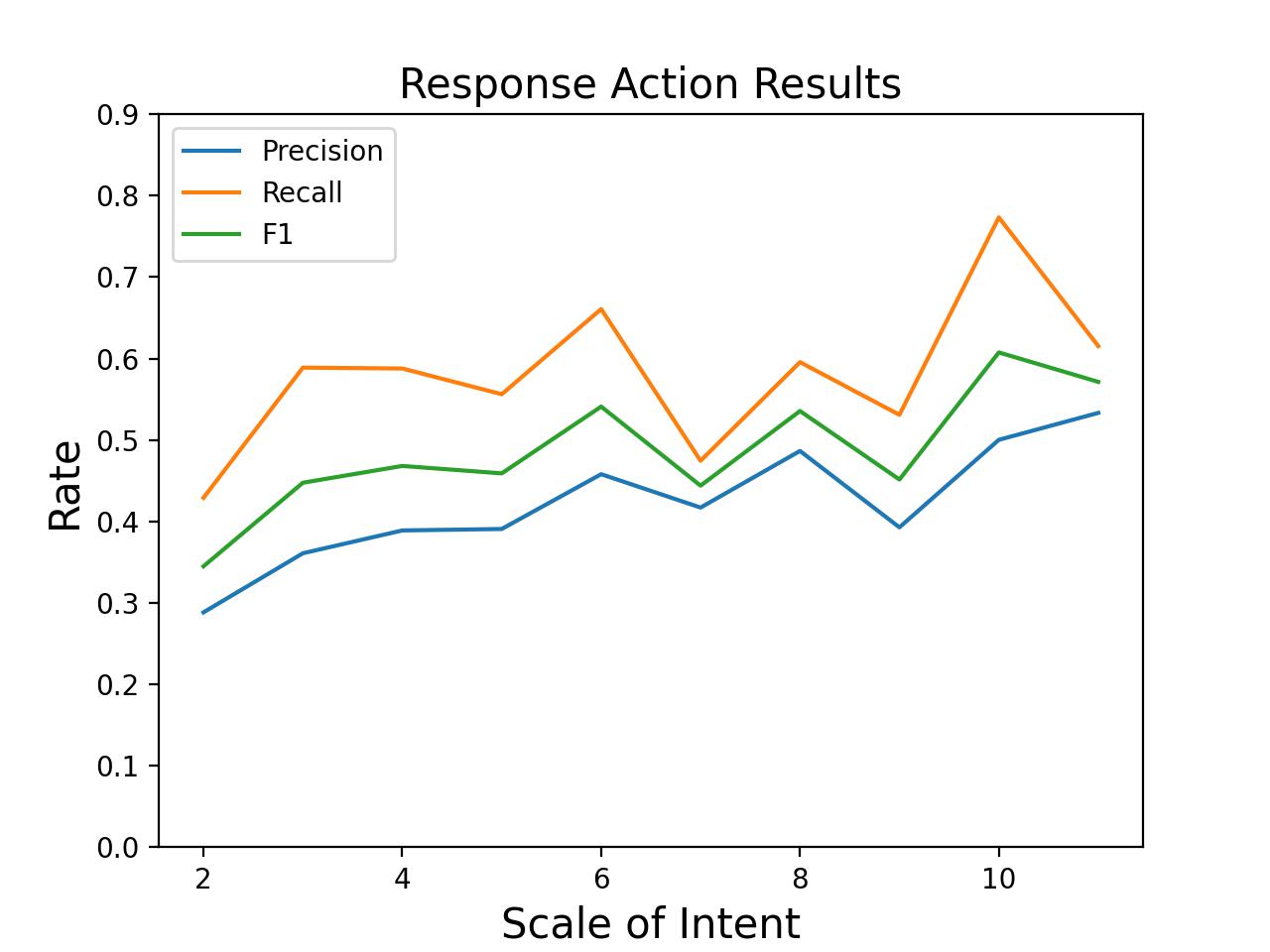}
	\label{figure:round_patience}
\end{minipage}
}
\caption{The Results of PUS Ablation}
\label{figure:slot_acc}
\end{figure}
\label{sec:experiment}

\begin{figure*}[htbp]
\centering
\includegraphics[height = 0.8\columnwidth]{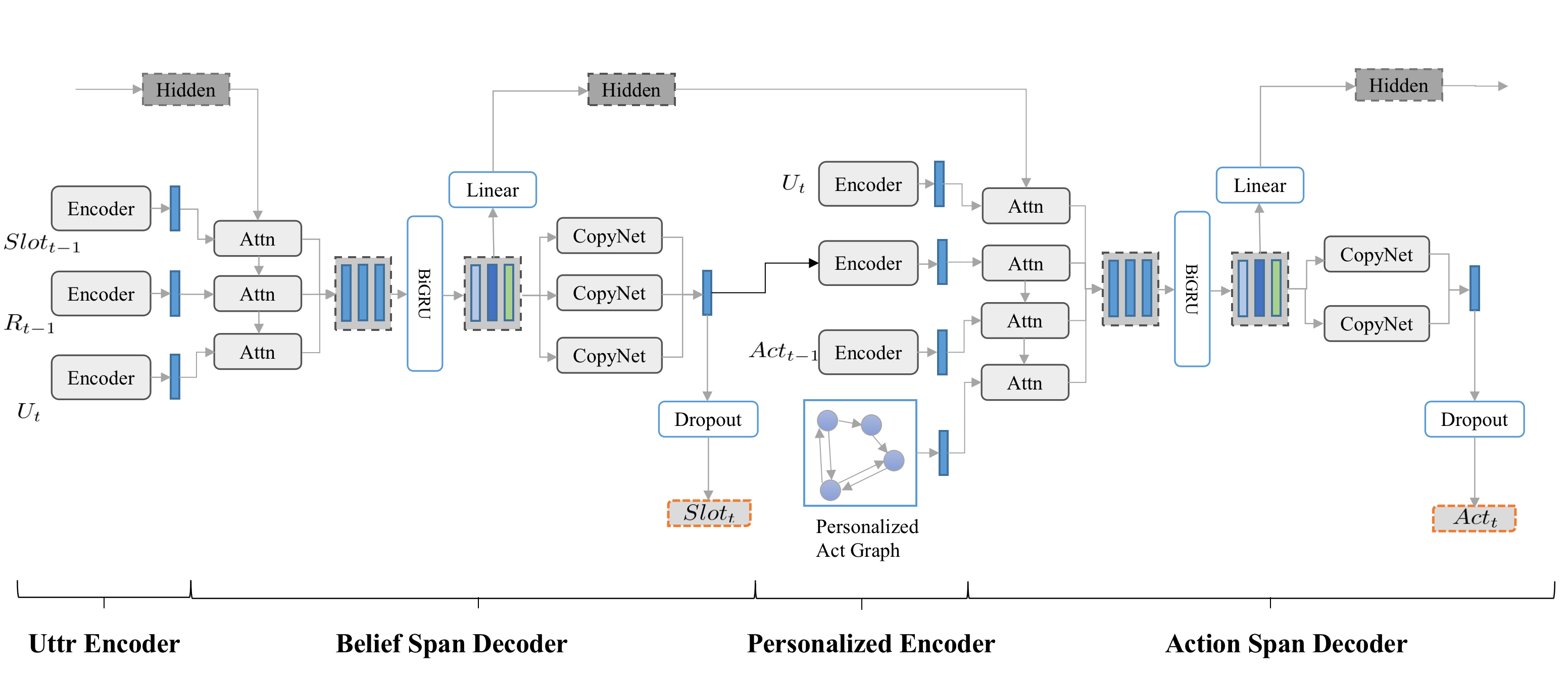}
\caption{Structure of DAMD Net with PUS embedding}
\label{figure:net_frame}
\end{figure*}
% 对于需求获取任务，由于不需要对实体服务的推荐，对话系统更加依赖于对用户的 了解才能做出有效响应。
% 虽然在真实数据的系统动作选择中并不一定是最优的需求引导策略，但对话动作的准确性能够体现/反应出从对话历史中学习PUS的有效性。
% 对对话系统来说，选择与数据集更加匹配的响应动作意味着更加符合真实的对话效果。因此，响应动作的准确性也可以从对话问答对的角度反应出对于需求获取策略的精确性。
% 我们在重标注的 CrossWOZ 数据集上，实现对于面向需求获取 任务的有效性实验。同样利用 Convlab 框架，我们实现了 Agenda-based Simulator，他会 对目标意图树作状态更新，同时为了更好的模拟用户的对话风格，利用栈结构存储预料中的用户行为动作的顺序，在对话过程中根据系统反馈动作，更新栈中行为序列和意图 树状态。实验结果如下所示。

%% Mingyi: 尽管你在这说SRE并不需要推荐服务、但是这个的评估应该还是要依赖于下游的评估任务吧，你这是个新任务，那问题来了，这个任务真的对下游任务有有益处么，如果没有做这个任务的意义在哪，，个人感觉你的实验评估还是得想办法圆回来。
% \my{see comment}
% SRE的优势之一是可以实现用户和服务提供者的解耦, 双方不直接交互而是借助意图树匹配相应的需求
One of the advantages of SRE is that it enables decoupling of users and service providers, where the two parties do not need to interact but achieve to match the corresponding requirements with the assistance of $R\_Tree$.
% Since there is no need to recommend services in the SRE task, dialogue system relies more on the perception of user to choose response actions effectively. 
While system response actions on real data are not necessarily the optimal requirement guidance strategy, the accuracy (PRF) of response actions can demonstrate the effectiveness of PUS model from dialogue history.
We perform the PUS validity experiments of response action based on the Convlab framework \cite{zhu2020convlab}. Specifically, we implement an Agenda-based simulator and a dialogue policy incorporating the PUS model to interact in CrossWOZ. The dialogue policy is based on the rule policy in Convlab. We also choose worker id as the different user identifier. 
As a comparison, we extract action state transfer trends and requirement planning trends, called global tendency of dialog strategy, from all dialogue data. This global tendency contains the feature of generic response action.
% 作为对比，我们从所有用户的对话数据抽取动作状态转移趋势和意图顺序趋势，称为全局趋势。这种全局趋势包含了通用的对话动作选择特征。
The experiment results are shown in Fig. \ref{figure:slot_acc}.

% 结果图分别为我们使用全局风格与个性化对话风格实现的对话策略，在需求获取任 务上的表现。可以看到使用个性化的对话风格对于对话动作反馈的有效性有较大提升。 另外，得益于对话风格模型对远距离上下文的记忆，随着对话意图点数量增多，能够更 加依赖先前对话信息获得更有效的相应准确率。
The results show that the strategy with PUS has a significant improvement in the PRF of response action, which means the response action considering personalized features can get more effective selection in real scenarios. In addition, as the scale of requirement increases, dialogue strategy can rely more on the previous context information to get more effective response accuracy due to the distant contextual memory of utterance style model.

\subsubsection{The Validity of PUS for Deep Network}

It is necessary for deep learning models to have the ability to embed the user personalized feature, which is a potential preference hidden in user dialogue.
%PUS is a potential preference hidden in user dialogue, which theoretically can also provide necessary user personalized features for deep learning models. 
Therefore,  we designed a personalized feature extraction method, which can effectively vectorize the PUS model and integrate it into the deep learning model. 
%The auxiliary model has the ability to distinguish different user dialogue styles. ability. 
% Specifically, for the important components in dialogue style: action-state transition preference and intent-planning preference, we use wFST and FPMC models to predict the next action. 
% Specifically, the trained wFST can give the probability density vector of the transition from the current state to each possible dialogue action. We use this vector as the state embedding, which contains the user's hidden preference information for all possible types of actions. For the trained FPMC model, the probability density of all possible intentions in the next step of the intention point of the current context is also given. Considering that the next intention selection information is constrained in the leaf nodes of the intention tree, we perform mask operation on the intention points in the intention tree that have already appeared. For the two probability density vectors, we use a linear layer for mapping and connection respectively.
The PUS can be embedded with following equation: % \ref{eq:pus_embed}
\begin{equation}
\begin{split}
        PUS_{embed}=&Concat(MLP(wFST(u,DA_{seq})),\\             
        &MASK(MLP(FPMC(u,Req_{seq}))))
\end{split}
\label{eq:pus_embed}
\end{equation}
Taking DAMD\cite{zhang2020task} model as an example, which is an end-to-end task dialogue model. It tackles multi-domain response generation problem through proposed multi-action data augmentation framework.
% The encoder is responsible for encoding the text information combined with the context, and the three-stage decoders are responsible for decoding information at three levels: slot identification, action decision-making and text generation. However, the original DAMD model did not distinguish user roles to user personalized information. Therefore 
We try to incorporate PUS embedding into DAMD structure of action decision-making step. The whole structure is shown in the Fig. \ref{figure:net_frame}. Note that we ignore the text generation module of original DAMD because it is not necessary for SRE task.

\begin{table}

\caption{Results on DAMD with PUS}
\label{table:damd}
\centering
\small
\begin{tabular}{ccc} 
\toprule
                         & \textbf{DAMD} & \textbf{DAMD (PUS)}  \\ 
\hline
\textbf{Joint Goal (\%)} & 46.2          & \textbf{47.6}           \\
\textbf{Slot Acc (\%)}   & 96.3          & \textbf{96.4}           \\
\textbf{Slot F1 (\%)}    & 88.3          & \textbf{88.7}           \\
% \textbf{Act F1 (\%)}     & 49.3          & \textbf{50.3}           \\
% \textbf{Match (\%)}      & \textbf{89.9} & 81.6                    \\
% \textbf{Success (\%)}    & \textbf{74.8} & 64.5                    \\
\bottomrule
\end{tabular}

\end{table}

The results in the Table \ref{table:damd} 
show that the DAMD model incorporating PUS features can improve the effect of Slot and Act prediction.
However, the DAMD is an end-to-end dialog model and model whole process of dialog. We notice that the PUS only works on fine-grained utterance, while for overall dialogue indicators, such as dialogue target matching (Match), 
there is a large gap in performance. % of the dialogue success rate (Success). 
The possible reason is that we introduced PUS features in dialogue action recognition and decision-making module. 
% And we set up a special loss function to assist dialogue action and slot convergence, so the global performance of dialog may be decreased due to the imbalance of dialog action data.
% However, the reverse conduction of the loss of actions and slots will also affect the gradients of other modules, which will have a greater impact on the overall model. Better results should be obtained with efficient tuning of the model's hyperparameters.

\section{Conclusion}
\label{sec:conclusion}
In this work, we focus on a human-machine dialogue system for service requirement elicitation task.
% 本文主要研究了一种面向需求获取任务的人机对话系统。
We design a dialogue strategy base on the personalized utterance style(PUS) model and requirement pattern to effectively elicit user requirements.
% 针对需求获取任务的两个特点：用户个性化的依赖性以及意图表达的随机性，我们分别使用个性化对话风格模型和基于需求模式的意图树模型作处理。
% Unlike general user profiles, PUS tries to model user's potential tendencies of requirement expression.
%不同于传统的用户画像，PUS用来建模用户在表达意图时的潜在倾向。
The dialogue strategy is able to dynamically expand the requirement tree and search target requirement based on cross-domain requirement patterns.
% 需求获取策略基于跨领域的需求模式，动态拓展对话意图树，实现有针对性的需求引导。
We validated effectiveness of the dialogue strategy incorporating PUS model in a customized dataset, which can achieve the SRE task efficiently.
% 在定制的对话数据集当中，我们验证了融合对话风格的对话策略的有效性。
As an essential part of service computing, efficient service requirement elicitation is the solid foundation for subsequent service provision.
% Dialogue systems can provide requirement-based services in more user-friendly ways.
In the future, we hope to explore more factors of personalized utterance style in complex dialogue strategies to achieve a better experience for the service environment.

\section{Conflict of Interest Statement}

All authors declare that they have no conflicts of interest.

\bibliography{ref}% common bib file


\begin{thebibliography}{26}
\ifx \bisbn   \undefined \def \bisbn  #1{ISBN #1}\fi
\ifx \binits  \undefined \def \binits#1{#1}\fi
\ifx \bauthor  \undefined \def \bauthor#1{#1}\fi
\ifx \batitle  \undefined \def \batitle#1{#1}\fi
\ifx \bjtitle  \undefined \def \bjtitle#1{#1}\fi
\ifx \bvolume  \undefined \def \bvolume#1{\textbf{#1}}\fi
\ifx \byear  \undefined \def \byear#1{#1}\fi
\ifx \bissue  \undefined \def \bissue#1{#1}\fi
\ifx \bfpage  \undefined \def \bfpage#1{#1}\fi
\ifx \blpage  \undefined \def \blpage #1{#1}\fi
\ifx \burl  \undefined \def \burl#1{\textsf{#1}}\fi
\ifx \doiurl  \undefined \def \doiurl#1{\url{https://doi.org/#1}}\fi
\ifx \betal  \undefined \def \betal{\textit{et al.}}\fi
\ifx \binstitute  \undefined \def \binstitute#1{#1}\fi
\ifx \binstitutionaled  \undefined \def \binstitutionaled#1{#1}\fi
\ifx \bctitle  \undefined \def \bctitle#1{#1}\fi
\ifx \beditor  \undefined \def \beditor#1{#1}\fi
\ifx \bpublisher  \undefined \def \bpublisher#1{#1}\fi
\ifx \bbtitle  \undefined \def \bbtitle#1{#1}\fi
\ifx \bedition  \undefined \def \bedition#1{#1}\fi
\ifx \bseriesno  \undefined \def \bseriesno#1{#1}\fi
\ifx \blocation  \undefined \def \blocation#1{#1}\fi
\ifx \bsertitle  \undefined \def \bsertitle#1{#1}\fi
\ifx \bsnm \undefined \def \bsnm#1{#1}\fi
\ifx \bsuffix \undefined \def \bsuffix#1{#1}\fi
\ifx \bparticle \undefined \def \bparticle#1{#1}\fi
\ifx \barticle \undefined \def \barticle#1{#1}\fi
\bibcommenthead
\ifx \bconfdate \undefined \def \bconfdate #1{#1}\fi
\ifx \botherref \undefined \def \botherref #1{#1}\fi
\ifx \url \undefined \def \url#1{\textsf{#1}}\fi
\ifx \bchapter \undefined \def \bchapter#1{#1}\fi
\ifx \bbook \undefined \def \bbook#1{#1}\fi
\ifx \bcomment \undefined \def \bcomment#1{#1}\fi
\ifx \oauthor \undefined \def \oauthor#1{#1}\fi
\ifx \citeauthoryear \undefined \def \citeauthoryear#1{#1}\fi
\ifx \endbibitem  \undefined \def \endbibitem {}\fi
\ifx \bconflocation  \undefined \def \bconflocation#1{#1}\fi
\ifx \arxivurl  \undefined \def \arxivurl#1{\textsf{#1}}\fi
\csname PreBibitemsHook\endcsname

%%% 1
\bibitem{yu2021incorporating}
\begin{bchapter}
\bauthor{\bsnm{Yu}, \binits{D.}},
\bauthor{\bsnm{Tian}, \binits{J.}},
\bauthor{\bsnm{Su}, \binits{T.}},
\bauthor{\bsnm{Tu}, \binits{Z.}},
\bauthor{\bsnm{Xu}, \binits{X.}},
\bauthor{\bsnm{Wang}, \binits{Z.}}:
\bctitle{Incorporating multimodal sentiments into conversational bots for
  service requirement elicitation}.
In: \bbtitle{IEEE International Conference on Service-Oriented System
  Engineering},
pp. \bfpage{81}--\blpage{90}
(\byear{2021}).
\bcomment{IEEE}
\end{bchapter}
\endbibitem

%%% 2
\bibitem{tian2022intention}
\begin{botherref}
\oauthor{\bsnm{Tian}, \binits{J.}},
\oauthor{\bsnm{Tu}, \binits{Z.}},
\oauthor{\bsnm{Li}, \binits{N.}},
\oauthor{\bsnm{Su}, \binits{T.}},
\oauthor{\bsnm{Xu}, \binits{X.}},
\oauthor{\bsnm{Wang}, \binits{Z.}}:
Intention model based multi-round dialogue strategies for conversational ai
  bots.
Applied Intelligence,
1--25
(2022)
\end{botherref}
\endbibitem

%%% 3
\bibitem{xu2020domain}
\begin{bchapter}
\bauthor{\bsnm{Xu}, \binits{H.}},
\bauthor{\bsnm{Wang}, \binits{X.}},
\bauthor{\bsnm{Wang}, \binits{Y.}},
\bauthor{\bsnm{Li}, \binits{N.}},
\bauthor{\bsnm{Tu}, \binits{Z.}},
\bauthor{\bsnm{Wang}, \binits{Z.}},
\bauthor{\bsnm{Xu}, \binits{X.}}:
\bctitle{Domain priori knowledge based integrated solution design for internet
  of services}.
In: \bbtitle{2020 IEEE International Conference on Services Computing (SCC)},
pp. \bfpage{446}--\blpage{453}
(\byear{2020}).
\bcomment{IEEE}
\end{bchapter}
\endbibitem

%%% 4
\bibitem{mazare2018training}
\begin{bchapter}
\bauthor{\bsnm{Mazare}, \binits{P.-E.}},
\bauthor{\bsnm{Humeau}, \binits{S.}},
\bauthor{\bsnm{Raison}, \binits{M.}},
\bauthor{\bsnm{Bordes}, \binits{A.}}:
\bctitle{Training millions of personalized dialogue agents}.
In: \bbtitle{Proceedings of the 2018 Conference on Empirical Methods in Natural
  Language Processing},
pp. \bfpage{2775}--\blpage{2779}
(\byear{2018})
\end{bchapter}
\endbibitem

%%% 5
\bibitem{zhu2020crosswoz}
\begin{botherref}
\oauthor{\bsnm{Zhu}, \binits{Q.}},
\oauthor{\bsnm{Huang}, \binits{K.}},
\oauthor{\bsnm{Zhang}, \binits{Z.}},
\oauthor{\bsnm{Zhu}, \binits{X.}},
\oauthor{\bsnm{Huang}, \binits{M.}}:
Cross{WOZ}: A large-scale chinese cross-domain task-oriented dialogue dataset.
Transactions of the Association for Computational Linguistics
(2020)
\end{botherref}
\endbibitem

%%% 6
\bibitem{lee2019structural}
\begin{barticle}
\bauthor{\bsnm{Lee}, \binits{C.-H.}},
\bauthor{\bsnm{Chen}, \binits{C.-H.}},
\bauthor{\bsnm{Trappey}, \binits{A.J.}}:
\batitle{A structural service innovation approach for designing smart product
  service systems: Case study of smart beauty service}.
\bjtitle{Advanced Engineering Informatics}
\bvolume{40},
\bfpage{154}--\blpage{167}
(\byear{2019})
\end{barticle}
\endbibitem

%%% 7
\bibitem{xiang2007srem}
\begin{bchapter}
\bauthor{\bsnm{Xiang}, \binits{J.}},
\bauthor{\bsnm{Liu}, \binits{L.}},
\bauthor{\bsnm{Qiao}, \binits{W.}},
\bauthor{\bsnm{Yang}, \binits{J.}}:
\bctitle{Srem: A service requirements elicitation mechanism based on ontology}.
In: \bbtitle{31st Annual International Computer Software and Applications
  Conference (COMPSAC 2007)},
vol. \bseriesno{1},
pp. \bfpage{196}--\blpage{203}
(\byear{2007}).
\bcomment{IEEE}
\end{bchapter}
\endbibitem

%%% 8
\bibitem{hwang2014service}
\begin{barticle}
\bauthor{\bsnm{Hwang}, \binits{S.-Y.}},
\bauthor{\bsnm{Hsu}, \binits{C.-C.}},
\bauthor{\bsnm{Lee}, \binits{C.-H.}}:
\batitle{Service selection for web services with probabilistic qos}.
\bjtitle{IEEE transactions on services computing}
\bvolume{8}(\bissue{3}),
\bfpage{467}--\blpage{480}
(\byear{2014})
\end{barticle}
\endbibitem

%%% 9
\bibitem{wang2021graph}
\begin{barticle}
\bauthor{\bsnm{Wang}, \binits{Z.}},
\bauthor{\bsnm{Chen}, \binits{C.-H.}},
\bauthor{\bsnm{Zheng}, \binits{P.}},
\bauthor{\bsnm{Li}, \binits{X.}},
\bauthor{\bsnm{Khoo}, \binits{L.P.}}:
\batitle{A graph-based context-aware requirement elicitation approach in smart
  product-service systems}.
\bjtitle{International Journal of Production Research}
\bvolume{59}(\bissue{2}),
\bfpage{635}--\blpage{651}
(\byear{2021})
\end{barticle}
\endbibitem

%%% 10
\bibitem{lee2019sumbt}
\begin{bchapter}
\bauthor{\bsnm{Lee}, \binits{H.}},
\bauthor{\bsnm{Lee}, \binits{J.}},
\bauthor{\bsnm{Kim}, \binits{T.-Y.}}:
\bctitle{Sumbt: Slot-utterance matching for universal and scalable belief
  tracking}.
In: \bbtitle{the 57th Annual Meeting of the Association for Computational
  Linguistics},
pp. \bfpage{5478}--\blpage{5483}
(\byear{2019})
\end{bchapter}
\endbibitem

%%% 11
\bibitem{takanobu2019guided}
\begin{bchapter}
\bauthor{\bsnm{Takanobu}, \binits{R.}},
\bauthor{\bsnm{Zhu}, \binits{H.}},
\bauthor{\bsnm{Huang}, \binits{M.}}:
\bctitle{Guided dialog policy learning: Reward estimation for multi-domain
  task-oriented dialog}.
In: \bbtitle{the 2019 Conference on Empirical Methods in Natural Language
  Processing and the 9th International Joint Conference on Natural Language
  Processing},
pp. \bfpage{100}--\blpage{110}
(\byear{2019})
\end{bchapter}
\endbibitem

%%% 12
\bibitem{zhang2020task}
\begin{bchapter}
\bauthor{\bsnm{Zhang}, \binits{Y.}},
\bauthor{\bsnm{Ou}, \binits{Z.}},
\bauthor{\bsnm{Yu}, \binits{Z.}}:
\bctitle{Task-oriented dialog systems that consider multiple appropriate
  responses under the same context}.
In: \bbtitle{AAAI Conference on Artificial Intelligence},
vol. \bseriesno{34},
pp. \bfpage{9604}--\blpage{9611}
(\byear{2020})
\end{bchapter}
\endbibitem

%%% 13
\bibitem{qian2021learning}
\begin{bchapter}
\bauthor{\bsnm{Qian}, \binits{H.}},
\bauthor{\bsnm{Dou}, \binits{Z.}},
\bauthor{\bsnm{Zhu}, \binits{Y.}},
\bauthor{\bsnm{Ma}, \binits{Y.}},
\bauthor{\bsnm{Wen}, \binits{J.-R.}}:
\bctitle{Learning implicit user profile for personalized retrieval-based
  chatbot}.
In: \bbtitle{the 30th ACM International Conference on Information \& Knowledge
  Management},
pp. \bfpage{1467}--\blpage{1477}
(\byear{2021})
\end{bchapter}
\endbibitem

%%% 14
\bibitem{madotto-etal-2019-personalizing}
\begin{bchapter}
\bauthor{\bsnm{Madotto}, \binits{A.}},
\bauthor{\bsnm{Lin}, \binits{Z.}},
\bauthor{\bsnm{Wu}, \binits{C.-S.}},
\bauthor{\bsnm{Fung}, \binits{P.}}:
\bctitle{Personalizing dialogue agents via meta-learning}.
In: \bbtitle{the 57th Annual Meeting of the Association for Computational
  Linguistics},
pp. \bfpage{5454}--\blpage{5459}
(\byear{2019})
\end{bchapter}
\endbibitem

%%% 15
\bibitem{zhang2020memory}
\begin{barticle}
\bauthor{\bsnm{Zhang}, \binits{B.}},
\bauthor{\bsnm{Xu}, \binits{X.}},
\bauthor{\bsnm{Li}, \binits{X.}},
\bauthor{\bsnm{Ye}, \binits{Y.}},
\bauthor{\bsnm{Chen}, \binits{X.}},
\bauthor{\bsnm{Wang}, \binits{Z.}}:
\batitle{A memory network based end-to-end personalized task-oriented dialogue
  generation}.
\bjtitle{Knowledge-Based Systems}
\bvolume{207},
\bfpage{106398}
(\byear{2020})
\end{barticle}
\endbibitem

%%% 16
\bibitem{luo2019learning}
\begin{bchapter}
\bauthor{\bsnm{Luo}, \binits{L.}},
\bauthor{\bsnm{Huang}, \binits{W.}},
\bauthor{\bsnm{Zeng}, \binits{Q.}},
\bauthor{\bsnm{Nie}, \binits{Z.}},
\bauthor{\bsnm{Sun}, \binits{X.}}:
\bctitle{Learning personalized end-to-end goal-oriented dialog}.
In: \bbtitle{AAAI Conference on Artificial Intelligence},
vol. \bseriesno{33},
pp. \bfpage{6794}--\blpage{6801}
(\byear{2019})
\end{bchapter}
\endbibitem

%%% 17
\bibitem{he2021conversation}
\begin{bchapter}
\bauthor{\bsnm{He}, \binits{M.}},
\bauthor{\bsnm{Shen}, \binits{T.}},
\bauthor{\bsnm{Dong}, \binits{R.}}:
\bctitle{Conversation and recommendation: Knowledge-enhanced personalized
  dialog system}.
In: \bbtitle{International Conference on Web Engineering},
pp. \bfpage{209}--\blpage{224}
(\byear{2021}).
\bcomment{Springer}
\end{bchapter}
\endbibitem

%%% 18
\bibitem{chen2017survey}
\begin{barticle}
\bauthor{\bsnm{Chen}, \binits{H.}},
\bauthor{\bsnm{Liu}, \binits{X.}},
\bauthor{\bsnm{Yin}, \binits{D.}},
\bauthor{\bsnm{Tang}, \binits{J.}}:
\batitle{A survey on dialogue systems: Recent advances and new frontiers}.
\bjtitle{Acm Sigkdd Explorations Newsletter}
\bvolume{19}(\bissue{2}),
\bfpage{25}--\blpage{35}
(\byear{2017})
\end{barticle}
\endbibitem

%%% 19
\bibitem{10.1162/tacl_a_00420}
\begin{barticle}
\bauthor{\bsnm{Żelasko}, \binits{P.}},
\bauthor{\bsnm{Pappagari}, \binits{R.}},
\bauthor{\bsnm{Dehak}, \binits{N.}}:
\batitle{{What Helps Transformers Recognize Conversational Structure?
  Importance of Context, Punctuation, and Labels in Dialog Act Recognition}}.
\bjtitle{Transactions of the Association for Computational Linguistics}
\bvolume{9},
\bfpage{1163}--\blpage{1179}
(\byear{2021})
\end{barticle}
\endbibitem

%%% 20
\bibitem{zhou2010casia}
\begin{bchapter}
\bauthor{\bsnm{Zhou}, \binits{K.}},
\bauthor{\bsnm{Li}, \binits{A.}},
\bauthor{\bsnm{Yin}, \binits{Z.}},
\bauthor{\bsnm{Zong}, \binits{C.}}:
\bctitle{Casia-cassil: A chinese telephone conversation corpus in real
  scenarios with multi-leveled annotation}.
In: \bbtitle{Proceedings of the Seventh International Conference on Language
  Resources and Evaluation (LREC'10)}
(\byear{2010})
\end{bchapter}
\endbibitem

%%% 21
\bibitem{bunt2010towards}
\begin{bchapter}
\bauthor{\bsnm{Bunt}, \binits{H.}},
\bauthor{\bsnm{Alexandersson}, \binits{J.}},
\bauthor{\bsnm{Carletta}, \binits{J.}},
\bauthor{\bsnm{Choe}, \binits{J.-W.}},
\bauthor{\bsnm{Fang}, \binits{A.C.}},
\bauthor{\bsnm{Hasida}, \binits{K.}},
\bauthor{\bsnm{Lee}, \binits{K.}},
\bauthor{\bsnm{Petukhova}, \binits{V.}},
\bauthor{\bsnm{Popescu-Belis}, \binits{A.}},
\bauthor{\bsnm{Romary}, \binits{L.}}, \betal:
\bctitle{Towards an iso standard for dialogue act annotation}.
In: \bbtitle{Seventh Conference on International Language Resources and
  Evaluation (LREC'10)}
(\byear{2010})
\end{bchapter}
\endbibitem

%%% 22
\bibitem{Minliu}
\begin{bchapter}
\bauthor{\bsnm{Liu}, \binits{M.}},
\bauthor{\bsnm{Tu}, \binits{Z.}},
\bauthor{\bsnm{Xu}, \binits{X.}},
\bauthor{\bsnm{Wang}, \binits{Z.}}:
\bctitle{Dialogue-based continuous update of user portraits}.
In: \bbtitle{2021 IEEE International Conference on Services Computing (SCC)},
pp. \bfpage{193}--\blpage{202}
(\byear{2021}).
\doiurl{10.1109/SCC53864.2021.00032}
\end{bchapter}
\endbibitem

%%% 23
\bibitem{zhou_augmenting_2019}
\begin{bchapter}
\bauthor{\bsnm{Zhou}, \binits{Y.}},
\bauthor{\bsnm{Tsvetkov}, \binits{Y.}},
\bauthor{\bsnm{Black}, \binits{A.W.}},
\bauthor{\bsnm{Yu}, \binits{Z.}}:
\bctitle{Augmenting {Non}-{Collaborative} {Dialog} {Systems} with {Explicit}
  {Semantic} and {Strategic} {Dialog} {History}}.
In: \bbtitle{the 8th International Conference on Learning Representations},
pp. \bfpage{1}--\blpage{12}
(\byear{2019})
\end{bchapter}
\endbibitem

%%% 24
\bibitem{rendle2010factorizing}
\begin{bchapter}
\bauthor{\bsnm{Rendle}, \binits{S.}},
\bauthor{\bsnm{Freudenthaler}, \binits{C.}},
\bauthor{\bsnm{Schmidt-Thieme}, \binits{L.}}:
\bctitle{Factorizing personalized markov chains for next-basket
  recommendation}.
In: \bbtitle{the 19th International Conference on World Wide Web},
pp. \bfpage{811}--\blpage{820}
(\byear{2010})
\end{bchapter}
\endbibitem

%%% 25
\bibitem{schatzmann-etal-2007-agenda}
\begin{bchapter}
\bauthor{\bsnm{Schatzmann}, \binits{J.}},
\bauthor{\bsnm{Thomson}, \binits{B.}},
\bauthor{\bsnm{Weilhammer}, \binits{K.}},
\bauthor{\bsnm{Ye}, \binits{H.}},
\bauthor{\bsnm{Young}, \binits{S.}}:
\bctitle{Agenda-based user simulation for bootstrapping a {POMDP} dialogue
  system}.
In: \bbtitle{Human Language Technologies 2007: The Conference of the North
  {A}merican Chapter of the Association for Computational Linguistics},
pp. \bfpage{149}--\blpage{152}
(\byear{2007})
\end{bchapter}
\endbibitem

%%% 26
\bibitem{zhu2020convlab}
\begin{bchapter}
\bauthor{\bsnm{Zhu}, \binits{Q.}},
\bauthor{\bsnm{Zhang}, \binits{Z.}},
\bauthor{\bsnm{Fang}, \binits{Y.}},
\bauthor{\bsnm{Li}, \binits{X.}},
\bauthor{\bsnm{Takanobu}, \binits{R.}},
\bauthor{\bsnm{Li}, \binits{J.}},
\bauthor{\bsnm{Peng}, \binits{B.}},
\bauthor{\bsnm{Gao}, \binits{J.}},
\bauthor{\bsnm{Zhu}, \binits{X.}},
\bauthor{\bsnm{Huang}, \binits{M.}}:
\bctitle{Convlab-2: An open-source toolkit for building, evaluating, and
  diagnosing dialogue systems}.
In: \bbtitle{Proceedings of the 58th Annual Meeting of the Association for
  Computational Linguistics: System Demonstrations},
pp. \bfpage{142}--\blpage{149}
(\byear{2020})
\end{bchapter}
\endbibitem

\end{thebibliography}

\end{document}